\documentclass[11pt]{article}

\usepackage[preprint]{acl} % camera-ready with authors and page numbers
\usepackage{times}
\usepackage{latexsym}
\usepackage[T1]{fontenc}
\usepackage[utf8]{inputenc}
\usepackage{microtype}
\IfFileExists{inconsolata.sty}{\usepackage{inconsolata}}{}

\usepackage{amsmath}
\usepackage{amssymb}      % \varnothing and other AMS symbols (acmart used to supply these)
\usepackage{graphicx}
\usepackage{booktabs}
\usepackage{caption}
\usepackage{xcolor}       % also pulled in by tikz/colortbl, loaded first for safety
\usepackage{url}

\usepackage{xspace}
\newcommand{\ourmethod}{RIT-RAG\xspace}

\providecommand{\Description}[1]{}

\usepackage{stfloats}
\usepackage{adjustbox}

\usepackage{multirow}
\usepackage{tabularx}
\newcolumntype{Y}{>{\centering\arraybackslash}X}
\usepackage{colortbl}
\definecolor{famVanilla}{HTML}{F0F1F3}  % neutral gray
\definecolor{famAgentic}{HTML}{FBF0E4}  % amber
\definecolor{famGraph}{HTML}{EFEAF7}    % violet
\definecolor{famOurs}{HTML}{E3F1E9}     % green (ours)

\usepackage{algorithm}
\usepackage{algpseudocode}
\algrenewcommand{\algorithmiccomment}[1]{\hfill\(\triangleright\)~\parbox[t]{0.42\linewidth}{\raggedright #1}}

\usepackage{listings}
\usepackage[breakable,skins]{tcolorbox}
\definecolor{promptbg}{gray}{0.96}

\newtcolorbox{promptbox}{
  breakable,
  enhanced,
  colback=promptbg,
  colframe=gray!55,
  boxrule=0.5pt,
  arc=0pt,
  left=6pt, right=6pt, top=5pt, bottom=5pt,
  boxsep=0pt,
  before skip=6pt, after skip=6pt,
}

\newcommand{\phead}[1]{\par\noindent\textbf{#1}\par\nobreak\noindent\ignorespaces}

\usepackage{tikz}
\usetikzlibrary{arrows.meta,positioning,calc,fit,backgrounds,shapes.geometric}

\definecolor{cInk}{HTML}{22303F}       % near-black text / strong lines
\definecolor{cBlue}{HTML}{2F6FED}      % primary: retrieval / query path
\definecolor{cBlueDeep}{HTML}{1B4DB1}
\definecolor{cAmber}{HTML}{E8863A}     % agent
\definecolor{cGreen}{HTML}{16A37A}     % answer (success)
\definecolor{cGrey}{HTML}{8A94A6}      % soft grey lines / unexplored
\definecolor{plotRAG}{HTML}{0072B2}
\definecolor{plotAgentic}{HTML}{E69F00}
\definecolor{plotInteract}{HTML}{D55E00}
\definecolor{plotRIT}{HTML}{009E73}
\definecolor{plotHyperGraph}{HTML}{56B4E9}
\definecolor{plotCogRAG}{HTML}{CC79A7}
\definecolor{plotPageIndex}{HTML}{404040}
\definecolor{cFill}{HTML}{EDF0F5}      % neutral block fill
\definecolor{cGreyBG}{HTML}{F2F4F8}    % offline phase tint
\definecolor{cBlueBG}{HTML}{E9F1FE}    % online phase tint

\tikzset{
  block/.style={rounded corners=3pt, draw=cGrey, fill=cFill, line width=0.7pt,
                minimum height=8mm, align=center, inner sep=4pt, font=\small,
                text=cInk},
  ablock/.style={rounded corners=3pt, draw=cBlue, fill=cBlue!8, line width=1pt,
                 minimum height=8mm, align=center, inner sep=4pt, font=\small,
                 text=cInk},
  rblock/.style={rounded corners=3pt, draw=cAmber, fill=cAmber!12,
                 line width=1pt, minimum height=8mm, align=center,
                 inner sep=4pt, font=\small, text=cInk},
  gblock/.style={rounded corners=3pt, draw=cGreen, fill=cGreen!10,
                 line width=1pt, minimum height=8mm, align=center,
                 inner sep=4pt, font=\small, text=cInk},
  pill/.style={rounded corners=4pt, draw=cBlue, fill=cBlue!8,
               font=\footnotesize\ttfamily, inner sep=3.5pt, minimum width=31mm,
               align=center, text=cBlueDeep},
  gnode/.style={circle, draw=cGrey, fill=cFill, minimum size=4.4mm, inner sep=0},
  rnode/.style={circle, draw=cBlueDeep, fill=cBlue, line width=0.6pt,
                minimum size=5.4mm, inner sep=0, font=\scriptsize, text=white},
  pnode/.style={circle, draw=cBlue, fill=white, line width=1pt,
                minimum size=4.8mm, inner sep=0},
  arrow/.style={-{Latex[length=2.4mm]}, draw=cGrey, line width=0.8pt},
  aarrow/.style={-{Latex[length=2.4mm]}, draw=cBlue, line width=1.1pt},
  o2arrow/.style={-{Latex[length=2.4mm]}, draw=cAmber, line width=1.1pt},
  readarrow/.style={-{Latex[length=2.2mm]}, draw=cGrey, line width=0.8pt, dashed},
  litedge/.style={draw=cBlue, line width=1.3pt},
  greyedge/.style={draw=cGrey, line width=0.7pt},
  cyl/.style={cylinder, shape border rotate=90, aspect=0.25, draw=cGrey,
              fill=cFill, minimum width=13mm, minimum height=11mm, inner sep=2pt,
              font=\footnotesize, text=cInk, align=center},
  capt/.style={font=\footnotesize\itshape, text=cGrey},
  seclbl/.style={font=\small\bfseries},
}

\tikzset{
  tqbox/.style={rounded corners=3pt, draw=cInk, fill=white, line width=0.8pt,
                text width=42mm, align=center, font=\footnotesize, inner sep=4pt,
                text=cInk, minimum height=9mm},
  ttool/.style={rounded corners=4pt, draw=cBlue, fill=cBlue!8, font=\scriptsize\ttfamily,
                inner sep=4pt, text width=42mm, align=center, text=cBlueDeep,
                minimum height=7mm},
  ttoolR/.style={ttool, draw=cBlueDeep},
  tdist/.style={rounded corners=2pt, draw=cAmber, fill=cAmber!12, line width=0.8pt,
                text width=42mm, align=left, font=\scriptsize\ttfamily, inner sep=4pt,
                text=cInk},
  tgood/.style={tdist, draw=cGreen, fill=cGreen!10},
  tthink/.style={rounded corners=2pt, draw=cGrey, fill=cFill, line width=0.6pt,
                 text width=42mm, align=left, font=\scriptsize\itshape, inner sep=4pt,
                 text=cInk},
  tbad/.style={tthink, draw=cAmber, fill=cAmber!12},
  tans/.style={rounded corners=3pt, line width=1pt, text width=36mm, align=center,
               font=\footnotesize, inner sep=4pt, text=cInk, minimum height=8mm},
  tansbad/.style={tans, draw=red!70!black, fill=red!5},
  tansgood/.style={tans, draw=cGreen, fill=cGreen!10},
  tfaint/.style={text width=42mm, align=center, font=\scriptsize, text=cGrey,
                 inner sep=2pt},
  tflow/.style={-{Latex[length=2.8mm]}, draw=cGrey, line width=0.9pt},
  tart/.style={rounded corners=3pt, draw=cGrey, fill=white, line width=1.1pt,
               text width=52mm, align=left, font=\scriptsize\ttfamily, inner sep=5pt,
               text=cInk},
  tartbad/.style={tart, draw=cAmber},
  tartgood/.style={tart, draw=cGreen},
  thead/.style={align=center, font=\small\bfseries},
  tsub/.style={align=center, font=\scriptsize\itshape, text=cGrey},
}
\usepackage{pifont}
\usepackage{amssymb}
\usepackage{booktabs}
\usepackage{tcolorbox}           % no-op if already loaded (avoids option clash)
\tcbuselibrary{skins,raster}     % add the libraries we need on top

\providecommand{\ourmethod}{RIT-RAG\xspace}

\definecolor{oursC}{RGB}{0,105,120}      % our method (teal)
\definecolor{baseC}{RGB}{96,99,110}      % baselines (slate)
\definecolor{okC}{RGB}{22,128,60}        % correct
\definecolor{badC}{RGB}{196,32,44}       % error
\definecolor{okBG}{RGB}{225,244,230}
\definecolor{badBG}{RGB}{253,228,228}
\definecolor{chipBG}{RGB}{236,238,242}
\definecolor{mutedC}{RGB}{105,108,118}
\definecolor{stripBG}{RGB}{246,247,249}

\newcommand{\trOK}{\textcolor{okC}{\ding{51}}}
\newcommand{\trBAD}{\textcolor{badC}{\ding{55}}}

\newcommand{\trTool}[1]{\tcbox[on line, boxsep=0.6pt, left=1.6pt, right=1.6pt,
  top=0.4pt, bottom=0.4pt, arc=1.5pt, boxrule=0pt, colback=chipBG]{%
  \ttfamily\scriptsize #1}}
\newcommand{\trGood}[1]{\tcbox[on line, boxsep=0.5pt, left=1.2pt, right=1.2pt,
  top=0.3pt, bottom=0.3pt, arc=1pt, boxrule=0pt, colback=okBG]{\strut #1}}
\newcommand{\trBad}[1]{\tcbox[on line, boxsep=0.5pt, left=1.2pt, right=1.2pt,
  top=0.3pt, bottom=0.3pt, arc=1pt, boxrule=0pt, colback=badBG]{\strut #1}}
\newcommand{\trThought}[1]{{\color{mutedC}\itshape ``#1''}}
\newcommand{\trWhy}[1]{{\color{badC}\scriptsize\sffamily\bfseries $\blacktriangleright$\,#1}}
\newcommand{\trWhyOK}[1]{{\color{okC}\scriptsize\sffamily\bfseries $\blacktriangleright$\,#1}}
\newcommand{\trDoc}[1]{{\ttfamily\scriptsize #1}}
\newcommand{\trElide}{{\color{mutedC}\,[\ldots]\,}}
\newcommand{\trToc}[2]{\par\noindent\hspace*{\dimexpr1.1em+#1\dimexpr0.9em\relax\relax}%
  {\color{mutedC}$\llcorner$}\,{\scriptsize #2}}

\newcounter{tracestep}
\newcommand{\trStepNum}{\stepcounter{tracestep}%
  \tcbox[on line, boxsep=0pt, left=1.5pt, right=1.5pt, top=0.5pt, bottom=0.5pt,
  arc=4pt, boxrule=0pt, colback=black!70]{\color{white}\sffamily\bfseries\fontsize{5.5}{6}\selectfont\thetracestep}}
\newcommand{\trStep}[1]{\par\smallskip\noindent\makebox[1.1em][l]{\trStepNum}%
  \parbox[t]{\dimexpr\linewidth-1.1em\relax}{\raggedright #1\strut}\par}
\newcommand{\trSub}[1]{\par\noindent\hspace*{1.1em}%
  \parbox[t]{\dimexpr\linewidth-1.1em\relax}{\raggedright #1\strut}\par}

\newtcolorbox{qstrip}{enhanced, colback=stripBG, colframe=black!18, boxrule=0.4pt,
  arc=2pt, left=5pt, right=5pt, top=3pt, bottom=3pt, fontupper=\footnotesize}
\newcommand{\trQ}{{\sffamily\bfseries Q:}~}
\newcommand{\trGold}{{\sffamily\bfseries\color{okC}Gold:}~}

\tcbset{
  tracebase/.style={enhanced, arc=2pt, boxrule=0.6pt,
    left=4pt, right=4pt, top=3pt, bottom=3pt, middle=2.5pt,
    fonttitle=\sffamily\bfseries\small, fontupper=\footnotesize, fontlower=\footnotesize,
    toptitle=1pt, bottomtitle=1pt, segmentation style={solid, black!25, line width=0.4pt}},
  ours/.style={tracebase, colframe=oursC, colbacktitle=oursC, colback=white},
  base/.style={tracebase, colframe=baseC, colbacktitle=baseC, colback=white},
  okans/.style={colbacklower=okBG},
  badans/.style={colbacklower=badBG},
}
\newtcolorbox{tracebox}[4][base]{#1,
  title={#2\hfill{\normalfont\scriptsize #4}\hspace{4pt}%
    \tcbox[on line, boxsep=0pt, left=1.5pt, right=1.5pt, top=0pt, bottom=0pt,
      boxrule=0pt, arc=1.5pt, colback=white]{#3}},
  code={\setcounter{tracestep}{0}}}
\newcommand{\trAns}{{\sffamily\bfseries Answer:}~}
\newcommand{\trArgs}[1]{{\ttfamily\scriptsize\color{mutedC}(#1)}}
\newcommand{\trSep}{{\color{mutedC}\,\textbar\,}}

\usepackage{wrapfig}       % text-wrapped table of agent tools (figures/table_tools.tex)

\usepackage[normalem]{ulem} % \sout for \ydel
\newcommand{\eg}{\mbox{\it e.g.}}
\newcommand{\ie}{\mbox{\it i.e.}}

\newcommand{\viz}{\mbox{\it viz.}}

\newcommand{\corpus}{\mathcal{C}}
\newcommand{\numdocs}{M}

\newcommand{\doc}[1]{d_{#1}}
\newcommand{\toc}[1]{T_{#1}}

\newcommand{\tnode}{v}                              % a node of the tree
\newcommand{\tocnodes}[1]{V_{#1}}         % node set of the ToC tree of D_i
\newcommand{\content}{c_v}         % text span owned by a node
\newcommand{\rootpath}[1]{\mathrm{path}(#1)}        % nodes on the root-to-node path
\newcommand{\strrep}{\mathcal{S}}                 % string of a node or a tree
\newcommand{\chunk}{c}
\newcommand{\anc}[1]{\mathrm{node}(#1)}                    % node a chunk comes from
\newcommand{\query}{q}

\newcommand{\budget}{B}                             % max number of tool calls (hyperparameter)
\newcommand{\searchbudget}{B_{\mathrm{s}}}          % max number of Search calls (hyperparameter)
\newcommand{\selection}[1]{S_{#1}}                 % nodes selected for reading in round #1
\newcommand{\answer}{a}                            % final answer
\newcommand{\searchquery}{\tilde{q}}               % search query formulated by the agent
\newcommand{\score}[2]{s(#1, #2)}                  % score of node #1 for search query #2
\newcommand{\similarity}[2]{\mathrm{sim}(#1, #2)}  % similarity between a search query and a chunk
\newcommand{\retrieved}[2]{\mathcal{R}_{#1}(#2)}   % top-#1 chunks retrieved for query #2
\newcommand{\nonzeronodeset}[1]{\mathcal{N}_{#1}}  % nodes that contain a retrieved chunk
\newcommand{\numtopnodes}{n}                       % number of highest-scoring nodes kept by the Search tool
\newcommand{\expandednodeset}[1]{\mathcal{V}_{#1}} % those nodes plus all their ancestors
\newcommand{\qforest}[1]{\mathcal{F}_{#1}}         % forest induced by \expandednodeset{#1} (26 Sep: forest replaces the subtree)
\newcommand{\round}{t}                             % round index

\newcommand{\haiku}{Haiku\xspace}                       % generation-model shorthand
\newcommand{\haikofull}{Claude~Haiku~4.5\xspace}        % full model name (first mention, config table)
\newcommand{\gptmini}{GPT-5-mini\xspace}                % generation model
\newcommand{\luna}{Luna\xspace}                         % table/prose display name for GPT-5.6
\newcommand{\lunafull}{GPT-5.6-Luna\xspace}                  % underlying model id for Luna
\newcommand{\qwen}{Qwen3-32B\xspace}                    % open-source structure inducer / preprocessor
\newcommand{\embedder}{\texttt{gte-modernbert-base}\citep{zhang2024mgte}\xspace} % dense-retrieval embedder
\newcommand{\judgemodel}{Gemini~2.5~Flash\xspace}       % evaluation judge (binary + G-E)
\newcommand{\entdocs}{EntDocs\xspace}                  % name of the 2.84M-page enterprise documentation corpus (never say IBM)
\newcommand{\entdocsbench}{EntQABench\xspace}

\newcommand{\wixqap}[1]{%
  \ifcase#1\relax
    \or Pricing Plans: Canceling Client Plans%
    \or Pricing Plans: Making Site Pages Members-Only%
    \or Pricing Plans: Creating a Free Plan%
  \else\errmessage{wixqap: index #1 out of range (expected 1--3)}%
  \fi\xspace}

\newcommand{\wixqaa}[1]{%
  \ifcase#1\relax
    \or Account Settings Request: Viewing Login Activity for Your Account%
    \or Account Settings Request: Reusing an Account Name%
    \or Account Settings Request: Automatically Logging Out of Your Wix Account on All Your Devices%
  \else\errmessage{wixqaa: index #1 out of range (expected 1--3)}%
  \fi\xspace}

\newcommand{\gettoc}{\texttt{GetToC}}
\newcommand{\readtool}{\texttt{Read}}

\usepackage{graphicx}
\usepackage{fancyvrb}
\newsavebox{\jsonbox}

\usepackage{xcolor} % Required for coloring text

\begin{document}

% Title and authors.
% ---------------------------------------------------------------------------
% Front matter for ACL (ARR) format.
% \input by main.tex before \maketitle. The abstract lives in main.tex, after
% \maketitle, per the ICLR shell layout. ICLR has no CCS concepts or keywords.
% ---------------------------------------------------------------------------
\title{\texorpdfstring{\ourmethod}{RIT-RAG}: Navigating Document Corpora with Retrieval-Induced Trees}

\author{
  \normalsize\textbf{Meghanadh Pulivarthi\textsuperscript{1} \quad Swaraj Kumar Biswal\textsuperscript{2}\thanks{Work done during an internship at IBM.} \quad Kushagra Bhushan\textsuperscript{1}} \\
  \normalsize\textbf{Yatin Nandwani\textsuperscript{1} \quad Sachindra Joshi\textsuperscript{1} \quad Dinesh Raghu\textsuperscript{1}} \\
  \normalsize\textsuperscript{1}IBM \quad \textsuperscript{2}IIT Kharagpur \\
  \normalsize\texttt{\{Meghanadh.Pulivarthi1, kushagrabhushan, Yatin.Nandwani\}@ibm.com} \\
  \normalsize\texttt{\{jsachind, diraghu1\}@in.ibm.com} \\
  \normalsize\texttt{biswalswaraj88@kgpian.iitkgp.ac.in}
}

\maketitle

\begin{abstract}

Retrieval-augmented generation (RAG) grounds language models in external corpora. Agentic RAG enables iterative search, yet exposes the model to isolated chunks without document structure, making it difficult to distinguish relevant evidence from chunks that merely resemble the query. Structure-aware methods such as PageIndex navigate using the structure but cannot scale to structures of large corpora that cannot fit in the LLM context. Hence, they first commit to a single document using a document retriever and cannot recover from a wrong choice.
We propose \ourmethod (Retrieval-Induced Tree RAG), which combines content retrieval with structural navigation. Offline, \ourmethod builds a tree for each document from its table of contents or sitemap. At query time, it retrieves a broad set of chunks and uses their positions to induce manageable sub-trees, potentially across multiple documents. An LLM agent navigates these sub-trees, selectively reads promising nodes, and reformulates queries when needed. Thus, retrieval proposes where to look, while the agent decides what to read.
Across financial, scientific, and customer-support benchmarks, \ourmethod achieves the highest answer accuracy among vanilla, graph-based, and agentic baselines. On \entdocsbench, our new benchmark of 2.84 million technical-documentation webpages, it improves accuracy by 6.6--11.4 points over the strongest baseline across three LLMs.

\end{abstract}

\section{Introduction}

Retrieval-Augmented Generation (RAG) has emerged as a standard approach for grounding large language models (LLMs) in external knowledge~\citep{lewis2020rag,gao2024ragsurvey}. By retrieving relevant evidence from a corpus and providing it to an LLM, RAG enables models to answer questions using information beyond their parameters. It also supports citation and attribution. As applications increasingly require answering complex questions over large document collections, effective retrieval has become the central challenge.

Recent work has addressed this challenge through \textit{agentic RAG}, in which an LLM iteratively invokes search tools, examines the returned content, and reinvokes the search tools when necessary~\citep{yao2023react,asai2024selfrag,jeong2024adaptiverag,hui2026interactrag}. This enables the model to recover from unsuccessful retrieval attempts and perform multi-hop information seeking. However, agentic RAG operates primarily at the chunk level. Documents are split into independently searchable chunks, and the LLM reasons over the content of the chunks returned by the search tools. It has no explicit representation of how these chunks are organized within the document. Sections, subsections, and their relationships are therefore invisible to the retrieval process. 
As a result, without structural information, the agent struggles to distinguish the chunks that contain the answer (useful evidence) from those that are merely similar in terminology or topic (plausible evidence). 
%The resulting context contains more plausible evidence, but not necessarily more useful evidence, 
This leads to context overload due to unfocused retrieval.

A complementary line of work has explored \textit{structure-aware RAG}, exemplified by PageIndex~\citep{pageindex}. Rather than searching for content directly, these approaches first reason over a document's hierarchical structure, including sections, subsections, and their summaries, and then retrieve content from promising regions. This constitutes a top-down paradigm, proceeding from structure to content. However, the structure of an entire corpus cannot always fit in the context. The system must first select a document and then navigate its hierarchy. This creates an early commitment problem. Before seeing the document's content, the system has limited ability to compare competing documents. A wrong document selection can therefore lead to a coherent but irrelevant navigation path, with no easy way to recover.

These observations reveal a tension between retrieval recall and precision. Agentic RAG casts a wide net, achieving high recall: relevant evidence is often present somewhere among the retrieved chunks. However, precision is low, as the relevant signal is buried among loosely related content, making it difficult for the LLM to identify and use the right evidence. Structure-aware navigation presents the opposite trade-off. By following a narrow structural path, it can achieve high precision when the correct path is chosen. However, recall is limited: an incorrect choice of the document can prevent the LLM from ever seeing the relevant evidence.
%an incorrect choice at the document level can prevent the relevant evidence from ever being considered.

% \ycom{should we change it to Trees?}
We propose \textbf{\ourmethod} (Retrieval-Induced Tree RAG), a framework that addresses the precision--recall trade-off between agentic RAG and structure-aware RAG. \ourmethod begins with a conventional search
index, enabling scalable retrieval over large document collections. Rather than
directly exposing the retrieved chunks to the LLM, \ourmethod uses their metadata, \eg, the position of each chunk in the hierarchy of its document,
to
induce sub-trees representing the document structure around the retrieved chunks.
% \ycom{I would rather say:
% Rather than directly exposing the retrieved chunks to the LLM, \ourmethod uses their meta data to induce subtrees representing the structure around the documents containing the retrieved chunks.
% While i agree with kcom, but i am afraid that metadata of a chunk hasn't been introduced till now and it will be an overkill to do so. But for clarity and correctness, we should use metadata here.
% The issue with your framing is that you already know the methods section and experiments -- think about a reviewer who has only read the part of introduction that has come before this.
% }
The LLM is presented with this compact structural view and selects the
region most likely to contain the answer. It can then retrieve the contents of
that region. If no candidate region provides a promising path, the LLM
reformulates the query. The new query retrieves a different set of chunks and induces new sub-trees.
Retrieval and navigation, therefore, form an iterative loop.

This design enables \emph{selective context assembly}. Rather than dumping all retrieved chunks into the context, as in agentic RAG, or committing to a structural path before seeing relevant content, as in structure-aware RAG, 
%\ourmethod first uses content retrieval to anchor the search and then induces structure around the retrieved evidence. The resulting sub-trees provide the agent with a map of the relevant documents' neighborhoods, allowing it to selectively expand only the regions it considers promising. In this way, \ourmethod retains the broad coverage of content-based retrieval while exploiting document structure to focus subsequent retrieval.
\ourmethod first uses content retrieval to anchor the search and then exploits the existing structure surrounding the retrieved evidence. The resulting sub-trees provide the agent with the local structure surrounding the relevant passages, allowing it to selectively expand only the regions it considers promising. In this way, \ourmethod retains the broad coverage of content-based retrieval while exploiting document structure to guide subsequent retrieval.

The results show that \ourmethod consistently outperforms existing retrieval approaches across both public benchmarks and a large-scale enterprise setting. 
Across the three publicly available datasets (WixQA, FinanceBench, and QASPER), \ourmethod achieves the highest accuracy on all of them. 
It improves accuracy by 2.4--10.5 points over agentic RAG with the same retriever, and by 2.0--18.6 points over PageIndex,
demonstrating the benefit of combining content-based retrieval with structure-aware reasoning. The gains hold even when the native structure is too coarse: on FinanceBench, we induce finer tables of contents with an LLM, and \ourmethod still improves accuracy over the strongest baseline by 4.6--9.3 points. To evaluate scalability in a realistic enterprise setting, we introduce \entdocsbench, a new large-scale benchmark whose corpus, \entdocs, contains 2.84 million webpages from the product documentation of a large enterprise. On this benchmark, where the structure of \entdocs is too large for PageIndex to navigate, \ourmethod improves accuracy over the strongest baseline by 6.6--11.4 points, demonstrating that its advantages persist even at substantially larger corpus scales.
Our contributions are as follows.
\begin{itemize}
    \item We propose \ourmethod, a framework that combines content-based retrieval with structure-aware navigation. \ourmethod uses retrieved evidence to induce query-specific sub-trees, possibly from multiple documents,
    % Yatin: potentially encompassing multiple documents (i would prefer a better word for encompassing)
    enabling an LLM to reason over document structure before selectively retrieving the contents of promising regions.
    \item We construct the \entdocsbench, a new benchmark over \entdocs, a corpus of 2.84 million webpages from the product documentation of a large enterprise. It enables us to evaluate RAG systems at a realistic corpus scale. We will release the benchmark, including its corpus and questions, together with the document trees of all our corpora.
    \item We evaluate \ourmethod across multiple models and datasets and show that it consistently improves accuracy over existing RAG systems on three public benchmarks and the \entdocsbench. 
    %Besides agentic RAG methods, we compare against state-of-the-art non-agentic graph-based RAG methods, HyperGraphRAG~\citep{luo2025hypergraphrag} and Cog-RAG~\citep{hu2026cograg}. 
    An analysis of the text that each method reads confirms the precision--recall trade-off: agentic RAG has a high recall but a low precision, PageIndex has the opposite, and \ourmethod balances the two.
\end{itemize}
% \ycom{passages or chunks? what have we used in the introduction till now?}
\section{Related Work}

\paragraph{Retrieval-Augmented Generation}
(RAG) grounds language models in external knowledge by retrieving relevant information from a corpus before generation \citep{lewis2020rag}. Early RAG systems typically retrieve a fixed set of passages using lexical or dense similarity \citep{robertson2009bm25,karpukhin2020dpr} and provide them directly to the language model. \textbf{Agentic RAG} instead allows the model to iteratively reason about what information is needed and issue additional retrieval operations, with approaches including retrieval interleaved with reasoning \citep{trivedi2023ircot}, adaptive retrieval \citep{asai2024selfrag,jiang2023flare}, iterative retrieval-generation \citep{shao2023iterretgen} and hybrid retrieval techniques involving multiple tools the agent can use to retrieve content \citep{hui2026interactrag}. In all these settings, the model interacts with the corpus only through retrieval, and it never sees how the retrieved chunks are organized.
%within their documents.

\paragraph{Structured Retrieval:}
A limitation of passage-level retrieval is that independently ranking chunks can discard relationships among pieces of evidence. This has motivated retrieval methods that exploit additional structure in the corpus. One direction constructs \textbf{tree-based representations} of the corpus. RAPTOR \citep{sarthi2024raptor} recursively clusters and summarizes passages to construct a tree of increasingly abstract representations, while SiReRAG \citep{zhang2025sirerag} indexes similar and related information to support multihop reasoning. DocsRay \citep{jeong2025docsray} generates a pseudo table of contents for a document and uses it for hierarchical retrieval. Another direction constructs \textbf{graph-based representations} of the corpus. GraphRAG \citep{edge2024graphrag} builds an entity graph and community summaries to support retrieval and synthesis over related information, and HippoRAG \citep{gutierrez2024hipporag} runs personalized PageRank over a knowledge graph to retrieve passages. HyperGraphRAG \citep{luo2025hypergraphrag} extends this formulation to hypergraphs, enabling the representation of higher-order relationships among multiple entities or chunks. Cog-RAG \citep{hu2026cograg} further constructs complementary theme and entity hypergraphs, using thematic structure for coarse retrieval before fine-grained entity retrieval. A third direction structures the retrieved content at \textbf{inference time}. StructRAG \citep{li2024structrag} dynamically transforms retrieved information into data structures suited to different reasoning tasks using an LLM. In these approaches, however, the solver model remains oblivious to the underlying structure, and the retriever controls what the model reads rather than the model navigating the structure itself. Moreover, these structures are built by clustering, summarizing, or extracting entities, rather than 
exploiting the corpus's existing organization.
%exploiting the organization already present in the corpus.
%taken from the organization that the corpus already has.

\paragraph{Structure-Aware Agents:}
A recent line of work instead gives language models access to structure-guided navigation, allowing them to reason about where information is located and decide which parts of the structure to read. PageIndex \citep{pageindex} constructs a hierarchical table of contents for each document and presents the tree of the selected document to the model during inference. The model can then exploit its reasoning capabilities to navigate the tree and read only the sections it deems relevant. PDFTriage \citep{saadfalcon2024pdftriage} similarly gives the model tools to fetch the sections, pages, and tables of a document. These approaches navigate one document at a time. Over a large corpus, they must first select a document, typically with a retriever, and commit to it before seeing its content.

\ourmethod avoids this early commitment by combining content-based retrieval with structure-aware navigation. It retrieves from the entire corpus and shows the agent only the subtrees of the existing hierarchy around the retrieved chunks, which may span several documents. The agent then decides what to read, which separates evidence discovery from content reading.

% Method: setup, global ToC, retrieval-induced ToC, \ourmethod, pipeline figure.

\section{Preliminaries}
\label{sec:background}

\noindent \textbf{Corpus Representation}: Let $\corpus=\{\doc{1}, \doc{2}, \ldots, \doc{\numdocs}\}$ denote a corpus containing $\numdocs$ documents. Let $\toc{i}$ denote the hierarchical tree structure associated with document $\doc{i}$. Let $\tocnodes{i}$ be the set of nodes in $\toc{i}$. These nodes correspond to semantically organised units of the document, such as chapters, sections, and subsections, with parent-child relationships capturing their hierarchical organisation. We represent the corpus by the forest $\{\toc{1}, \ldots, \toc{\numdocs}\}$.

Each node $\tnode \in \tocnodes{i}$ owns the text that lies under it but not under any of its children, \eg, the introduction of a chapter before its first section. We denote this text by $\content$. Note that $\content$ is empty when all the text of a node lies under its children, \eg, a chapter that begins directly with its first section. Thus, every piece of text in $\doc{i}$ belongs to exactly one node. In addition, each node has associated metadata. The metadata contains a title, a summary of $\content$, and keywords present in $\content$. The title is mandatory, whereas the summary and keywords are optional: for a very large corpus containing millions of documents, it may not be feasible to extract them.
%and depend primarily on the instructions provided for extracting them and the content available within the node.

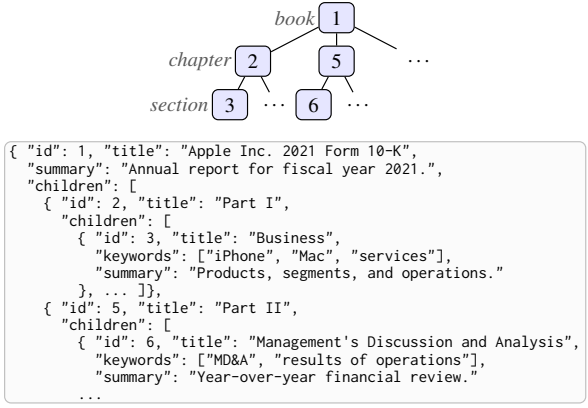
\begin{figure}[t]
\centering
\begin{lrbox}{\jsonbox}%
\begin{BVerbatim}[fontsize=\fontsize{5.8pt}{5.9pt}\selectfont]
{ "id": 1, "title": "Apple Inc. 2021 Form 10-K",
  "summary": "Annual report for fiscal year 2021.",
  "children": [
    { "id": 2, "title": "Part I",
      "children": [
        { "id": 3, "title": "Business",
          "keywords": ["iPhone", "Mac", "services"],
          "summary": "Products, segments, and operations."
        }, ... ]},
    { "id": 5, "title": "Part II",
      "children": [
        { "id": 6, "title": "Management's Discussion and Analysis",
          "keywords": ["MD&A", "results of operations"],
          "summary": "Year-over-year financial review."
        ...
\end{BVerbatim}
\end{lrbox}%
\resizebox{\columnwidth}{!}{%
\begin{tabular}{@{}c@{}}
\begin{tikzpicture}[
  every node/.style={font=\scriptsize},
  tn/.style={
    draw,
    rounded corners=2pt,
    inner sep=1.2pt,
    minimum width=4.2mm,
    minimum height=3.6mm
  },
  sm/.style={tn, fill=gray!15},
  dc/.style={tn, fill=blue!10},
  lbl/.style={
    font=\scriptsize\itshape,
    text=black!65,
    inner sep=1pt
  },
  level distance=5.2mm,
  sibling distance=9mm,
  level 1/.style={sibling distance=10mm},
  level 2/.style={sibling distance=5.5mm},
  edge from parent/.style={draw, -}
]
\node[dc, label={[lbl]left:book}] {1}
  child {
    node[dc, label={[lbl]left:chapter}] {2}
      child {node[dc, label={[lbl]left:section}] {3}}
      child {node[draw=none] {\dots}}
  }
  child {
    node[dc] {5}
      child {node[dc] {6}}
      child {node[draw=none] {\dots}}
  }
  child {node[draw=none] {\dots}};
\end{tikzpicture}\\[3pt]
\tcbox[
  colback=gray!3,
  colframe=gray!50,
  boxrule=0.4pt,
  arc=2pt,
  boxsep=0pt,
  left=1.5pt,
  right=1.5pt,
  top=1pt,
  bottom=1pt,
  nobeforeafter
]{\usebox{\jsonbox}}
\end{tabular}%
}
\caption{An illustrative example of a hierarchical tree structure associated
with a document and its JSON string representation.}
\label{fig:tree_string}
\end{figure}

Let $\strrep$ represent the serialization function that converts the structure of the tree $\toc{i}$ into a string $\strrep(\toc{i})$. This representation is consumed by the LLM, and it contains just the document's hierarchical structure without exposing the full content of each node. An example is shown in Figure~\ref{fig:tree_string}.
In $\strrep(\toc{i})$, each node is a JSON object with a unique identifier, its metadata, and the list of its children in their original order. The LLM uses the identifier to request a node's content.
% preamble: \usepackage{graphicx,fancyvrb}  and  \newsavebox{\jsonbox}

The tree structure may be explicitly available or extracted from the documents. For example, a book can be a document whose hierarchy can be derived from its table of contents and section headings. A group of webpages can be treated as a \textit{virtual document} whose hierarchical structure can be inferred from a sitemap or other metadata.

\noindent \textbf{Agentic Search}: Standard vector-index based RAG approaches split each document $\doc{i}$ into manageable chunks, embed each chunk using an off-the-shelf embedding model, such as BGE \citep{bgeembeddings}, and store the chunk embeddings in a vector index, \eg, using FAISS \citep{faiss} or OpenSearch \citep{opensearch}. During inference, based on its understanding of the user utterance and the ongoing conversation, the agent formulates a search query $\searchquery$, and the retriever retrieves the top $k$ chunks, $\retrieved{k}{\searchquery}$.
%with the highest retrieval scores for $\searchquery$.
The agent reads the retrieved chunks and decides to either generate the answer $\answer$, or formulate another search query to retrieve more chunks. 

%While this iterative process allows the agent to refine its search, the chunk-level index does not preserve the document's hierarchical structure.
%Consequently, the agent cannot directly leverage structural cues, such as the relationships among sections and subsections, while generating the answer.
%In addition, other than formulating queries, the agent has no control over what it reads, and it reads every retrieved chunk.
While this iterative process allows the agent to refine its search, chunk-level indexing discards the document's hierarchical structure.
Consequently, the agent cannot directly leverage structural cues, such as the hierarchical relationships between sections and subsections, when generating the answer.
In addition, apart from formulating queries, the agent has no control over which retrieved chunks it reads and must process all of them.

\noindent \textbf{PageIndex}: Given a user question that requires information from the corpus $\corpus$, the system first uses a document retriever to identify a document $\doc{i}$ that may contain the answer. The LLM is then provided with the string representation $\strrep(\toc{i})$, which encodes the hierarchical structure of document $\doc{i} $ without exposing its contents. The LLM navigates this structure to identify promising nodes and retrieves a node's content only when needed. It can iteratively inspect additional nodes until it has sufficient evidence to answer the question. Thus, unlike agentic search, the LLM has access to the document's structure up front while selectively accessing its contents.

This approach, however, has two limitations. First, since the structures of all documents cannot fit within the LLM's context, the system must first select a document to navigate. This early commitment relies solely on retrieval scores, and the agent cannot switch documents once committed, potentially leading to an incorrect path. Second, the tree representation can become prohibitively large even for a single large (virtual) document, such as webpage collections. When the full structure exceeds the context window, it must be truncated or partitioned, weakening the structural information available to the LLM.

\section{RIT-RAG: Retrieval-Induced Tree RAG}
\label{sec:ourmethod}

% Overview of RIT-RAG (Approach section). Drawn from the notation of the Approach section.
% Needs TikZ libraries arrows.meta, positioning, calc, fit, backgrounds (loaded in preamble.tex).
% Fonts: \footnotesize (8pt) for all labels and the table, \small bold for panel titles.
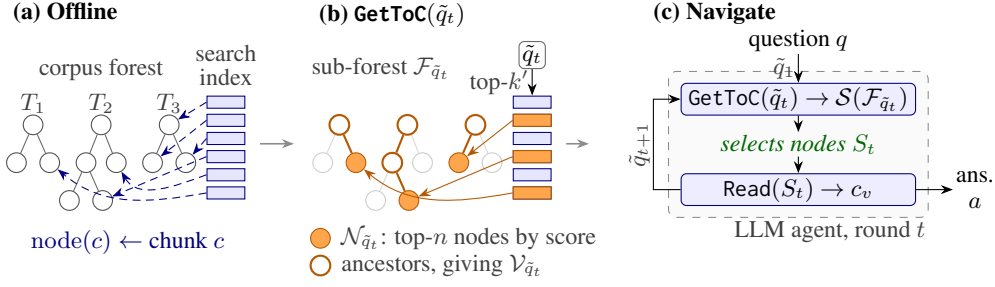
\begin{figure*}[t]
\centering
\begin{tikzpicture}[
  font=\footnotesize,
  >={Stealth[length=1.6mm]},
  tn/.style={circle, draw=black!70, fill=white, minimum size=2.6mm, inner sep=0pt, line width=0.4pt},
  hit/.style={fill=orange!70, draw=orange!70!black},
  anc/.style={draw=orange!70!black, line width=0.8pt},
  faded/.style={draw=black!20, fill=white},
  efaded/.style={draw=black!15},
  eanc/.style={draw=orange!70!black, line width=0.8pt},
  chunk/.style={draw=blue!50!black, fill=blue!10, minimum width=5mm, minimum height=1.6mm, inner sep=0pt},
  chunkhit/.style={chunk, fill=orange!70, draw=orange!70!black},
  box/.style={draw=black!60, rounded corners=2pt, fill=white, align=center, inner sep=2pt},
  panel/.style={font=\small\bfseries, anchor=north west},
  note/.style={font=\footnotesize, text=black!75, align=center},
  % A small corpus forest of three document trees (roots A, B, C). Every node and edge takes a style key,
  % so each panel can colour it differently.
  s0/.style={}, sA/.style={}, sB/.style={}, sC/.style={}, sa1/.style={}, sa2/.style={}, sb1/.style={},
  sb2/.style={}, sc1/.style={}, sc2/.style={}, sb11/.style={}, sb12/.style={},
  eA/.style={}, eB/.style={}, eC/.style={}, ea1/.style={}, ea2/.style={}, eb1/.style={},
  eb2/.style={}, ec1/.style={}, ec2/.style={}, eb11/.style={}, eb12/.style={},
  pics/gtree/.style={code={
    \node[tn,sA]   (-A)   at (-0.9,-0.5)  {};
    \node[tn,sB]   (-B)   at (0,-0.5)     {};
    \node[tn,sC]   (-C)   at (0.9,-0.5)   {};
    \node[tn,sa1]  (-a1)  at (-1.12,-1.0) {};
    \node[tn,sa2]  (-a2)  at (-0.68,-1.0) {};
    \node[tn,sb1]  (-b1)  at (-0.2,-1.0)  {};
    \node[tn,sb2]  (-b2)  at (0.2,-1.0)   {};
    \node[tn,sc1]  (-c1)  at (0.68,-1.0)  {};
    \node[tn,sc2]  (-c2)  at (1.12,-1.0)  {};
    \node[tn,sb11] (-b11) at (-0.42,-1.5) {};
    \node[tn,sb12] (-b12) at (0.02,-1.5)  {};
    \begin{scope}[on background layer]
      \draw[black!60,ea1]  (-A)--(-a1);  \draw[black!60,ea2]  (-A)--(-a2);
      \draw[black!60,eb1]  (-B)--(-b1);  \draw[black!60,eb2]  (-B)--(-b2);
      \draw[black!60,ec1]  (-C)--(-c1);  \draw[black!60,ec2]  (-C)--(-c2);
      \draw[black!60,eb11] (-b1)--(-b11); \draw[black!60,eb12] (-b1)--(-b12);
    \end{scope}
  }},
]

% ---------------- (a) Offline ----------------
\node[panel] at (-1.3,1.25) {(a) Offline};
\pic (T) at (0,0) {gtree};
\node[note, anchor=south] at ([yshift=3.2mm]T-B.north) {corpus forest};
\foreach \n/\i in {A/1, B/2, C/3} { \node[note, anchor=south, inner sep=0.5pt] at (T-\n.north) {$\toc{\i}$}; }
\foreach \i in {0,...,5} { \node[chunk] (ch\i) at (1.65,-0.2-0.24*\i) {}; }
\node[note, anchor=south, align=center] at (ch0.north) {search\\[-1pt]index};
\draw[->, blue!50!black, densely dashed] (ch1.west) -- (T-c1);
\draw[->, blue!50!black, densely dashed] (ch3.west) -- (T-b12);
\draw[->, blue!50!black, densely dashed] (ch5.west) to[out=190, in=-40] (T-a2);
\draw[->, blue!50!black, densely dashed] (ch0.west) -- (T-C);
\draw[->, blue!50!black, densely dashed] (ch2.west) -- (T-c2);
\draw[->, blue!50!black, densely dashed] (ch4.west) to[out=200, in=-60] (T-b2);
\node[note, text=blue!50!black, anchor=north] at (0.35,-1.75) {$\anc{\chunk} \leftarrow$ chunk $\chunk$};

% ---------------- (b) GetToC ----------------
\begin{scope}[xshift=4.05cm]
\node[panel] at (-1.3,1.25) {(b) \gettoc{}$(\searchquery_{\round})$};
\pic[
  sA/.style=anc, sB/.style=anc, sC/.style=anc, sb1/.style=anc,
  sa2/.style=hit, sb12/.style=hit, sc1/.style=hit,
  sa1/.style=faded, sb2/.style=faded, sc2/.style=faded, sb11/.style=faded,
  eb1/.style=eanc, ea2/.style=eanc, eb12/.style=eanc, ec1/.style=eanc,
  ea1/.style=efaded, eb2/.style=efaded, ec2/.style=efaded, eb11/.style=efaded
] (S) at (0,0) {gtree};
\node[note, anchor=south] at ([xshift=-3.5mm,yshift=3.2mm]S-B.north) {sub-forest $\qforest{\searchquery_{\round}}$};
\foreach \i in {0,...,5} { \node[chunk] (sh\i) at (1.65,-0.2-0.24*\i) {}; }
\node[chunkhit] at (sh1) {}; \node[chunkhit] at (sh3) {}; \node[chunkhit] at (sh5) {};
\node[box, anchor=south, inner sep=1.2pt] (sq) at ([yshift=3.5mm]sh0.north) {$\searchquery_{\round}$};
\draw[->] (sq) -- node[note, left, inner sep=1pt] {top-$k'$} (sh0.north);
\draw[->, orange!70!black] (sh1.west) -- (S-c1);
\draw[->, orange!70!black] (sh3.west) -- (S-b12);
\draw[->, orange!70!black] (sh5.west) to[out=190, in=-40] (S-a2);
% legend
\node[tn,hit] (L1) at (-1.15,-2.0) {}; \node[note, anchor=west] at (L1.east) {$\nonzeronodeset{\searchquery_{\round}}$: top-$\numtopnodes$ nodes by score};
\node[tn,anc] (L2) at (-1.15,-2.36) {}; \node[note, anchor=west] at (L2.east) {ancestors, giving $\expandednodeset{\searchquery_{\round}}$};
\end{scope}

% ---------------- (c) Navigate ----------------
\begin{scope}[xshift=8.12cm, stp/.style={box, minimum width=31mm, minimum height=4.2mm, inner sep=1pt},
              call/.style={stp, fill=blue!7, draw=blue!50!black}]
\node[panel] at (-0.95,1.25) {(c) Navigate};
\node[inner sep=1pt] (q) at (1.1,0.62) {question $\query$};
\node[call, below=3.8mm of q]  (s2) {\gettoc{}$(\searchquery_{\round})$ $\rightarrow$ $\strrep(\qforest{\searchquery_{\round}})$};
\node[below=2.2mm of s2, inner sep=0.8pt, font=\footnotesize\itshape, text=green!40!black] (s3) {selects nodes $\selection{\round}$};
\node[call, below=2.2mm of s3] (s4) {\readtool{}$(\selection{\round})$ $\rightarrow$ $\content$};
\begin{scope}[on background layer]
  \node[draw=black!45, dashed, rounded corners=3pt, fill=black!2, inner xsep=1.6mm, inner ysep=1.5mm, fit=(s2)(s4)] (loop) {};
\end{scope}
\node[note, anchor=north east, inner sep=1pt] (lbl) at (loop.south east) {LLM agent, round $\round$};
\draw[->] (q) -- node[note, left, inner sep=1pt] {$\searchquery_1$} (s2);
\draw[->] (s2) -- (s3); \draw[->] (s3) -- (s4);
\draw[->] (s4.west) -- ++(-3.6mm,0) |- node[note, pos=0.25, left, inner sep=1pt, rotate=90, anchor=south] {$\searchquery_{\round+1}$} (s2.west);
\node[inner sep=1pt, anchor=west, align=center] (ans) at ([xshift=3.2mm]loop.east |- s4) {ans.\\$\answer$};
\draw[->] (s4.east) -- (ans.west);
\end{scope}

% arrows between panels
\draw[->, black!45, line width=0.6pt] (2.1,-0.75) -- (2.55,-0.75);
\draw[->, black!45, line width=0.6pt] (6.15,-0.75) -- (6.45,-0.75);
\end{tikzpicture}
\caption{Overview of \ourmethod. (a) Offline, we index the text of the document trees $\toc{i}$, with each chunk $\chunk$ belonging to one node $\anc{\chunk}$. (b) Each query $\searchquery_{\round}$ retrieves top-$k'$ chunks; their top $\numtopnodes$ nodes and ancestors induce a sub-forest $\qforest{\searchquery_{\round}}$. (c) In each round, the LLM calls \gettoc{}, reasons over $\strrep(\qforest{\searchquery_{\round}})$, reads selected nodes $\selection{\round}$, and either answers or continues.}
\label{fig:overview}
\end{figure*}
Our approach, RIT-RAG, combines content-based retrieval with structure-aware navigation. We first construct a conventional search index over document chunks, as in agentic search. In addition to the chunk content, however, we associate each indexed chunk with lightweight structural metadata, including its node identifier, title, and the identifiers and titles of its ancestor nodes in the document tree. This metadata allows the hierarchical structure surrounding a retrieved chunk to be reconstructed at inference time.

Given a user question, the retriever returns the top-$k' > k$ chunks, $\retrieved{k'}{\searchquery}$, based on their retrieval scores. Rather than exposing their contents directly to the LLM, we use their structural metadata to reconstruct the corresponding subtrees and provide them as context. The LLM then reasons over these structures to identify promising nodes and selectively request their contents. Thus, structural context is induced by content retrieval, avoiding the need to inspect every document in the corpus.

This design addresses the limitations of both chunk-based agentic RAG and structure-first approaches such as PageIndex. Unlike conventional agentic RAG, the LLM can exploit hierarchical relationships among retrieved nodes when deciding what to retrieve next. At the same time, unlike structure-first approaches, the system does not need to expose or fit the complete structure of the corpus into the LLM context, nor does it need to commit to a single document before considering content-level evidence. Because the initial retrieval operates over the entire corpus, relevant structures from multiple documents can be surfaced simultaneously. The size of the structure presented to the LLM is therefore bounded by the number of selected nodes and their ancestors, rather than by the size of the corpus.

% Node Scoring paragraph below is superseded (25 Sep): the score now lives in Inducing Subtree (eq:node_score).
% \noindent \textbf{Node Scoring:}
% We first retrieve the top $k' \gg k$ chunks, $\retrieved{k'}{\searchquery}$, using the vector index.
% For each retrieved chunk $\chunk$, we identify the node $\anc{\chunk}$ to which it belongs, and use the similarity score $\similarity{\searchquery}{\chunk}$ returned by the index to assign a score to $\anc{\chunk}$.
% While chunking, we ensure that a chunk does not span multiple nodes.
% However, a node often spans multiple chunks, especially when it contains a lot of text.
% If multiple chunks map to the same node, we aggregate their similarity scores to get the final score of the node.
% Formally, the score of a node $\tnode$ is
% \begin{equation}
% \score{\tnode}{\searchquery} = \sum_{\chunk \in \retrieved{k'}{\searchquery},\ \anc{\chunk} = \tnode} \similarity{\searchquery}{\chunk}.
% \label{eq:node_score}
% \end{equation}

\noindent \textbf{Inducing a Sub-Forest:} We first retrieve the top-$k' > k$ chunks, $\retrieved{k'}{\searchquery}$, using the search index. Note that we chunk the text $\content$ of every node separately, so a chunk never spans multiple nodes.
However, a node often spans multiple chunks, especially when it contains a lot of text. Thus, each retrieved chunk $\chunk$ belongs to exactly one node, which we denote by $\anc{\chunk}$.
We score each node using the highest retrieval score among its retrieved chunks.
%returned by the index.
Formally, the score of a node $\tnode$ is
\vspace{-2pt}
\begin{equation}
\score{\tnode}{\searchquery} = \max_{\substack{\chunk \in \retrieved{k'}{\searchquery} \\ \anc{\chunk} = \tnode}} \similarity{\searchquery}{\chunk}.
\label{eq:node_score}
\end{equation}
\vspace{-2pt}
Let $\nonzeronodeset{\searchquery}$ be the set of the $\numtopnodes$ nodes with the highest scores, where $\numtopnodes$ is a hyperparameter.
To construct a query-dependent sub-forest, we expand $\nonzeronodeset{\searchquery}$ with the ancestors of the nodes in it, \ie, all the nodes that lie on the path from any node in $\nonzeronodeset{\searchquery}$ to the root of its tree.
%Let $\rootpath{\tnode}$ denote the set of nodes on the path from the root $\troot$ to a node $\tnode$.
Let $\rootpath{\tnode}$ denote the set of nodes on the path from $\tnode$ to the root of the tree to which it belongs.
Formally,
%\begin{equation}
$\expandednodeset{\searchquery} = \bigcup_{\tnode \in \nonzeronodeset{\searchquery}} \rootpath{\tnode}$.
%\label{eq:expanded_nodes}
%\end{equation}

The nodes in $\expandednodeset{\searchquery}$ induce a sub-forest $\qforest{\searchquery}$, with one sub-tree for every document $\doc{i}$ that contains a node in $\nonzeronodeset{\searchquery}$. The corresponding sub-tree is rooted at the root of $\toc{i}$.
% We represent $\subtree{\searchquery}$ as a string as described above, with one addition: each node $\tnode \in \nonzeronodeset{\searchquery}$ also shows its score $\score{\tnode}{\searchquery}$ in brackets after its title.
We represent $\qforest{\searchquery}$ as a string $\strrep(\qforest{\searchquery})$ by concatenating the string representations of its sub-trees. We list the sub-trees in decreasing order of their score, \ie, the highest score $\score{\tnode}{\searchquery}$ among the nodes $\tnode \in \nonzeronodeset{\searchquery}$ in the sub-tree.
Including the ancestors puts each retrieved node in a global context. For example, two sections titled ``Configuring the server'' may come from two versions of a product, and only their ancestors distinguish them. In addition, retrieved nodes that share an ancestor appear together under it, which shows the agent where the evidence concentrates.
% \ycom{This has to change. Maybe introduce strrep function for forest?}

% \input{figures/table_tools} % Table 1 (tools) dropped on 26 Sep: not adding much value.
\noindent \textbf{Navigating the Sub-Forest:}
In \ourmethod, the LLM agent works in rounds.
% \ycom{Instead of "As in tree-induced RAG", simply say: "In \ourmethod, the LLM agent works in rounds.}
In each round $\round$, the agent first formulates a search query $\searchquery_{\round}$, and calls the \gettoc{} tool to get $\strrep(\qforest{\searchquery_{\round}})$.
% \ycom{Lets come up with a better name than Search tool: maybe simply 'Get Structure' tool? Suggest something better. }
It then reasons over $\strrep(\qforest{\searchquery_{\round}})$ to select a set of nodes $\selection{\round}$, and reads their text using the \readtool{} tool.
Thus, in each round, we restrict the agent's search to the nodes in $\expandednodeset{\searchquery_{\round}}$.
Based on the contents, $\content$ for every $\tnode \in \selection{\round}$, it may generate the answer $\answer$, read more nodes from the same sub-forest $\qforest{\searchquery_{\round}}$, or formulate another search query for the next round. This is how \ourmethod recovers from an unpromising sub-forest. Since every \gettoc{} call retrieves from the entire corpus, a reformulated query can surface different documents, and the agent is never committed to the documents of an earlier round.
The agent stops when it generates the answer, or when it reaches a budget of $\budget$ tool calls, of which at most $\searchbudget$ can be calls to the \gettoc{} tool, where $\budget$ and $\searchbudget$ are hyperparameters.
Figure~\ref{fig:overview} illustrates this process.

\section{Experiments}
\label{sec:experiments}
Our experiments first evaluate how \ourmethod compares with existing baselines when navigating structure already present in a corpus (RQ1). Such a structure, however, is not always available: \eg, a PDF may lack a table of contents or provide one that is too coarse to support effective navigation. In such cases, the structure must first be induced. We therefore next evaluate how effectively \ourmethod navigates an induced structure (RQ2). Finally, we evaluate \ourmethod's scalability on a corpus of millions of interconnected webpages (RQ3).

\subsection{Experimental Setup}
\label{sec:setup}

\noindent \textbf{Datasets:} For RQ1, we use two datasets whose documents come with an existing structure.
WixQA \citep{cohen2025wixqa} has $200$ expert-written customer-support questions over $6{,}221$ help-center pages.
QASPER \citep{dasigi2021qasper} has $1{,}372$ answerable test questions over $416$ NLP papers.
Since its questions assume a known paper, we prefix each with ``According to the paper $\langle$title$\rangle$,'' and index all $1{,}585$ papers from all splits.
For RQ2, we use FinanceBench \citep{islam2023financebench}, which contains $150$ financial questions across $364$ SEC filings in PDF format.
For RQ3, we contribute \entdocsbench, a benchmark comprising the \entdocs corpus and $317$ synthetically generated questions covering three of its  products. The \entdocs corpus consists of $2.84$ million public webpages crawled from the product documentation of a large enterprise. See App.~\ref{sec:entdocs-creation} for details on \entdocsbench creation and App.~\ref{sec:stats} for statistics on all corpora.
%For RQ3, we contribute the \entdocsbench, whose corpus, \entdocs, consists of $2.84$ million public webpages crawled from the product documentation of a large enterprise, along with $317$ synthetically generated questions on three of its products.
%See App.~\ref{sec:entdocs-creation} for details on \entdocsbench creation and App.~\ref{sec:stats} for the statistics of all corpora.
%We will release the \entdocsbench and the document trees of all corpora.
%App.~\ref{sec:entdocs-creation} describes the generation, provenance tracking, and filtering of the \entdocsbench questions.
%App.~\ref{sec:stats} reports the statistics of all corpora.

\noindent \textbf{Building the Document Trees:}
We build the tree of each document from its table of contents (ToC), and the tree of a group of webpages from its sitemap (Section~\ref{sec:background}, details in App.~\ref{sec:corpus-extraction}).
%We build $\gtoc$ from the sitemap of each corpus and the tables of contents (ToCs) of its documents (Section~\ref{sec:ourmethod}, details in App.~\ref{sec:corpus-extraction}).
In WixQA, the HTML headings of a page give its ToC, and we build a sitemap by grouping the pages into products and the products into categories.
In QASPER, the sections of a paper give its ToC.
The native ToC of a FinanceBench filing is too coarse, so we induce a finer one with the PageIndex pipeline \citep{pageindex}, which parses the PDF and uses an LLM to extract its sections.
% \ycom{App. (FinanceBench): please add that the native ToC is itself read from the PDF with LLM calls, so it is noisy as well as coarse, and that our PageIndex pipeline parses the PDFs with pdfplumber instead of PyPDF2.}
QASPER and FinanceBench have no sitemap, so their corpora are forests of independent document trees.
In \entdocs, the breadcrumbs of the webpages give the sitemap, and each webpage is a single node.
Finally, an LLM generates the summary and keywords of every node, except in \entdocs, where the nodes carry only their titles because of the corpus size.
To generate summaries, we use \haiku for WixQA, and the open-source \qwen for QASPER and FinanceBench to reduce cost.

\noindent \textbf{Baselines:}
Vanilla RAG retrieves the top-$5$ chunks and answers with a single LLM call.
The graph-based methods, HyperGraphRAG \citep{luo2025hypergraphrag} and Cog-RAG \citep{hu2026cograg}, extract entities and hyperedges from the corpus offline.
For a question, they retrieve the relevant entities, hyperedges, and chunks, and answer from them with a single LLM call.
The agentic methods call the retriever as a tool in a loop.
Agentic Search is the agentic RAG baseline of sec.~\ref{sec:background}, and Interact-RAG \citep{hui2026interactrag} adds finer control over retrieval, \eg, entity matching.
PageIndex also uses the structure: its agent reasons over the full ToC of one document and reads the pages it selects.
To select the document, it relies on a document retriever. See App.~\ref{app:pageindex} for its implementation details.
% of the document retriever.  
The graph-based methods and PageIndex preprocess the corpus with the same LLM as \ourmethod.

\noindent \textbf{Models and Hyperparameters:} Every method generates the answer with three LLMs: \haikofull, \gptmini, and \lunafull (App.~\ref{sec:model-config}).
All methods share a hybrid retriever that fuses BM25 \citep{robertson2009bm25} and dense retrieval with \embedder and FAISS \citep{faiss}, over chunks of up to $512$ tokens (App.~\ref{app:rit-implementation}).
The \gettoc{} tool of \ourmethod retrieves the top-$k' = 200$ chunks and induces the sub-forest from the $\numtopnodes = 50$ highest-scoring nodes (Eq.~\ref{eq:node_score}).
We choose $\numtopnodes$ on held-out questions (App.~\ref{app:ablations}).
The agent makes at most $\budget = 10$ tool calls, of which at most $\searchbudget = 3$ are \gettoc{} calls.

\noindent \textbf{Evaluation:}
Our main metric is accuracy, judged by an LLM (\judgemodel) that compares the prediction with the gold answer.
We also measure the precision and recall of the passages a method reads against the gold evidence, as well as its preprocessing cost, API cost, and per-query latency.
In the App.~\ref{app:quality}, we report lexical token recall \citep{adlakha2024correctness} and the seven-dimensional generation evaluation (G-E) of HyperGraphRAG. %\citep{luo2025hypergraphrag}.

\noindent \textbf{Prompts:}
All LLM-based task prompts, including system messages for different baselines and the LLM-as-a-Judge, are available in App.~\ref{app:prompts}.
%\noindent \textbf{Prompts:} Prompts for all LLM-based tasks, including different system messages for various baselines, LLM-as-Judge used for evaluation, are available in App.~\ref{app:prompts}.
\subsection{Results}
\label{sec:results}

% One table covering WixQA + QASPER + FinanceBench with all three generation models (column order changed 25 Sep).
% One preprocessing model per corpus; IBM~Z moved to Table~\ref{tab:main-ibmz}.
\begin{table*}[t]
\centering\small
\setlength{\tabcolsep}{3pt}
\caption{Answer accuracy (\%) on WixQA, QASPER, and FinanceBench, by generation model. Best per column in \textbf{bold}.}
\label{tab:main}
\begin{tabular}{ll @{\hspace{7pt}} c @{\hspace{4pt}} c @{\hspace{4pt}} c @{\hspace{7pt}} c @{\hspace{4pt}} c @{\hspace{4pt}} c @{\hspace{7pt}} c @{\hspace{4pt}} c @{\hspace{4pt}} c}
\toprule
\multicolumn{2}{@{}l}{\textbf{Corpus}} & \multicolumn{3}{c}{WixQA} & \multicolumn{3}{c}{QASPER} & \multicolumn{3}{c}{FinanceBench} \\
\cmidrule(lr){3-5}\cmidrule(lr){6-8}\cmidrule(lr){9-11}
Family & Method & Haiku & \gptmini & Luna & Haiku & \gptmini & Luna & Haiku & \gptmini & Luna \\
\midrule
\addlinespace[1.5pt]
\rowcolor{famVanilla}\textit{Vanilla} & RAG ($k{=}5$) & 79.5 & 86.0 & 82.5 & 72.0 & 77.2 & 73.7 & 41.3 & 49.3 & 48.0 \\
\addlinespace[1.5pt]
\rowcolor{famGraph}\textit{Graph} & HyperGraphRAG & 87.5 & 94.0 & 87.5 & 44.8 & 52.6 & 52.6 & 29.3 & 35.3 & 47.3 \\
\rowcolor{famGraph} & Cog-RAG & 87.0 & 95.5 & 92.0 & 44.4 & 55.5 & 52.6 & 32.0 & 43.3 & 45.3 \\
\addlinespace[1.5pt]
\rowcolor{famAgentic}\textit{Agentic} & Agentic Search & 84.5 & 91.0 & 91.0 & 86.2 & 85.6 & 85.7 & 74.0 & 76.0 & 75.3 \\
\rowcolor{famAgentic} & Interact-RAG & 71.5 & 79.5 & 85.0 & 74.2 & 73.4 & 78.3 & 74.7 & 74.0 & 80.7 \\
\rowcolor{famAgentic} & PageIndex & 93.0 & 93.5 & 90.5 & 76.0 & 75.2 & 77.0 & 72.0 & 75.3 & 66.7 \\
\addlinespace[1.5pt]
\rowcolor{famOurs}\textit{Ours} & \ourmethod & \textbf{95.0} & \textbf{96.0} & \textbf{95.5} & \textbf{88.6} & \textbf{88.5} & \textbf{89.2} & \textbf{84.0} & \textbf{84.0} & \textbf{85.3} \\
\bottomrule
\end{tabular}
\end{table*}

\noindent \textbf{Performance on Datasets with Existing Structure (RQ1):}
Table~\ref{tab:main} reports the answer accuracy of all methods on WixQA, QASPER, and FinanceBench. 
\ourmethod achieves the highest accuracy across all three datasets and all three LLMs.
WixQA and QASPER already have an existing structure, and they let us test whether navigating it helps.
The most direct test is to compare \ourmethod with Agentic Search.
Both use the same retriever, the same LLM, and the same budget of $\budget = 10$ tool calls.
The main difference is what the agent sees after a search: Agentic Search sees the retrieved chunks, whereas \ourmethod sees the sub-forest induced by them.
This difference alone improves accuracy by $10.5$, $5.0$, and $4.5$ points on WixQA, and by $2.4$, $2.9$, and $3.5$ points on QASPER, with \haiku, \gptmini, and \luna, respectively.
PageIndex also uses the structure, but trails \ourmethod by $2.0$--$5.0$ points on WixQA and by $12.2$--$13.3$ points on QASPER.
It navigates the ToC of a single document, which we select by aggregating the retrieval scores of the chunks of each document (App.~\ref{app:pageindex}).
Once it selects a document, it cannot fetch the ToC of another one, and if the document is wrong, it cannot reach the evidence (Figure~\ref{fig:trace_lockheed}).
This is the early-commitment problem that we discussed earlier. 
\ourmethod does not commit to a document, since each search induces a new sub-forest that may span many documents.
The graph-based methods are competitive on WixQA, \eg, Cog-RAG reaches $95.5\%$ with \gptmini, but they fall far behind on QASPER, where they are worse than even vanilla RAG (App.~\ref{app:graph-failure}).
%App.~\ref{app:quality} reports lexical token recall and the G-E scores, which show similar trends.

\noindent \textbf{Performance on Datasets with Induced Structure (RQ2):}
On FinanceBench, \ourmethod navigates the ToCs induced by \qwen, and still improves over the best baseline by $9.3$, $8.0$, and $4.6$ points.
PageIndex navigates the same induced ToCs, but again trails \ourmethod, by $8.7$--$18.6$ points, as it commits to one retrieved document.

% Wrap table (25 Sep): Haiku-preprocessing accuracy, with the change w.r.t. Qwen3-32B preprocessing (Table tab:main) in brackets.
% Full two-block version: _review/tables_before_26sep/table_prep_sensitivity.tex
\begin{table}[t]
\centering\small
\setlength{\tabcolsep}{2.5pt}
\caption{Accuracy (\%) on FinanceBench when \haiku, instead of \qwen, induces the ToCs. In brackets: the change w.r.t.\ \qwen (Table~\ref{tab:main}).}
\label{tab:prep}
\vspace{-0.5ex}
\adjustbox{max width=\linewidth}{%
\begin{tabular}{ll ccc}
\toprule
Family & Method & Haiku & GPT-5 & Luna \\
\midrule
%\addlinespace[1.5pt]
\rowcolor{famVanilla}\textit{Vanilla} & RAG ($k{=}5$) & 44.7\,{\scriptsize(+3.4)} & 49.3\,{\scriptsize(\phantom{+}0.0)} & 46.0\,{\scriptsize($-$2.0)} \\
\addlinespace[1.5pt]
\rowcolor{famAgentic}\textit{Agentic} & Agentic Search & 76.7\,{\scriptsize(+2.7)} & 80.7\,{\scriptsize(+4.7)} & 78.7\,{\scriptsize(+3.4)} \\
\rowcolor{famAgentic} & Interact-RAG & 76.0\,{\scriptsize(+1.3)} & 76.0\,{\scriptsize(+2.0)} & 80.0\,{\scriptsize($-$0.7)} \\
\rowcolor{famAgentic} & PageIndex & 76.0\,{\scriptsize(+4.0)} & 77.3\,{\scriptsize(+2.0)} & 66.0\,{\scriptsize($-$0.7)} \\
\addlinespace[1.5pt]
\rowcolor{famOurs}\textit{Ours} & \ourmethod & \textbf{84.7}\,{\scriptsize(+0.7)} & \textbf{84.0}\,{\scriptsize(\phantom{+}0.0)} & \textbf{86.7}\,{\scriptsize(+1.4)} \\
\bottomrule
\end{tabular}%
}
\end{table}

\noindent \textbf{Quality of the Induced Structure:}
When we induce the structure, the accuracy of a structure-aware method may depend on the performance of the LLM that induces the structure.
To test this, we repeat the FinanceBench experiment with \haiku instead of \qwen as the LLM that induces the ToCs and generates the metadata (Table~\ref{tab:prep}). The accuracy of \ourmethod changes by at most $1.4$ points, showing that \ourmethod can effectively navigate structures induced by a small open-source LLM. In contrast, the baselines change by up to $4.7$ points, even though only PageIndex explicitly uses the induced ToCs. Recall that no chunk spans multiple nodes (Section~\ref{sec:ourmethod}), so the induced ToCs determine the chunk boundaries for all methods. Consequently, changing the induced ToCs alters the corpus chunking and, in turn, the chunks retrieved by the baselines. This makes the baselines sensitive to the induced structure, whereas \ourmethod is comparatively robust to such changes.

% Wrap table (25 Sep): \input it right before a paragraph. Needs wrapfig (loaded in custom_notations.tex).
\begin{table}[t]
\centering\small
\setlength{\tabcolsep}{3pt}
\caption{Answer accuracy (\%) on \entdocs, by generation model. Best per column in \textbf{bold}.}
\label{tab:main-ibmz}
\vspace{-0.5ex}
\adjustbox{max width=\linewidth}{%
\begin{tabular}{ll @{\hspace{6pt}} c @{\hspace{4pt}} c @{\hspace{4pt}} c}
\toprule
Family & Method & Haiku & GPT-5 & Luna \\
\midrule
\addlinespace[1.5pt]
\rowcolor{famVanilla}\textit{Vanilla} & RAG ($k{=}5$) & 43.2 & 57.4 & 52.1 \\
\addlinespace[1.5pt]
\rowcolor{famAgentic}\textit{Agentic} & Agentic Search & 73.5 & 75.4 & 77.0 \\
\rowcolor{famAgentic} & Interact-RAG & 74.8 & 76.0 & 77.6 \\
\rowcolor{famAgentic} & PageIndex & 57.4 & 63.1 & 69.4 \\
\addlinespace[1.5pt]
\rowcolor{famOurs}\textit{Ours} & \ourmethod & \textbf{81.4} & \textbf{83.9} & \textbf{89.0} \\
\bottomrule
\end{tabular}%
}
\end{table}

\noindent \textbf{Scaling to Millions of Documents (RQ3):}
Table~\ref{tab:main-ibmz} reports the accuracy on \entdocsbench. 
Note that the graph-based methods require LLM  calls over the entire corpus to construct their graphs, which becomes prohibitive at this scale (see sec.~\ref{sec:cost}). 
Consequently, we don't compare against them.
For PageIndex, we cluster webpages into virtual documents of manageable sizes. It retrieves the ToC of a single virtual document and reasons over it.
The large number of distractor (virtual) documents makes it harder for PageIndex to identify the relevant one, leading to lower end-to-end accuracy.
Recall that the nodes of \entdocs carry only their titles (Section~\ref{sec:setup}). Even so, \ourmethod improves over the best baseline by $6.6$, $7.9$, and $11.4$ points.
%Here, we compare only with methods that scale to $2.84$ million pages, \viz, vanilla RAG, Agentic Search, PageIndex and Interact-RAG. 
%The graph-based methods require LLM calls over the entire corpus to construct their graphs, which is prohibitive at this scale (Section~\ref{sec:cost}).
%\mycomment{Moreover, the large number of distractor documents in \entdocs makes it harder for PageIndex to identify the relevant documents, leading to lower end-to-end answer accuracy.} 

\subsection{Retrieval Analysis}
\label{sec:analysis}

To understand where the gains of \ourmethod come from, we examine what each method reads before answering. Table~\ref{tab:retrieval-haiku} reports the precision, recall, and F1 score of the passages read by each method with respect to the gold evidence on  WixQA, QASPER, and FinanceBench.
We exclude \entdocsbench from this analysis because multiple webpages may contain the correct evidence due to overlaps in the documentation of different product versions. However, only one webpage is annotated as gold. Consequently, an agent may retrieve valid evidence and answer correctly but still be penalized for not retrieving the annotated webpage.
We also exclude the graph-based methods because, in addition to passages, they read descriptions of the entities and hyperedges created during preprocessing, for which precision and recall cannot be measured.
%(App.~\ref{app:retrieval}).

Agentic Search calls the retriever multiple times and reads every chunk it retrieves. Consequently, it achieves high recall but low precision. PageIndex exhibits the opposite trend. It reads only the pages it selects from a single document, thereby achieving high precision. However, if it selects the wrong document, it cannot recover, causing its recall to drop, \eg, to $50.4\%$ on QASPER.
%, and to only $5.5\%$ on \entdocs, where the document it commits to rarely holds the evidence (App.~\ref{app:retrieval}). 
\ourmethod balances the two. It achieves the highest precision and recall on FinanceBench and the highest recall on WixQA, where only PageIndex is more precise, and ranks second on both metrics on QASPER. Consequently, it achieves the highest F1 on all three datasets.

These results support our motivation: the retriever allows the agent to search broadly, while the induced sub-forest allows it to read selectively. 
The analysis above uses \haiku as the LLM; we observe similar trends with the other LLMs.
(App.~\ref{app:retrieval}).
See App.~\ref{app:correct-overlap} for an alternative analysis that compares the sets of questions answered correctly by different methods and reaches the same conclusion. 
% requires \usepackage{array,graphicx}
\begin{table}[t]
\centering
\setlength{\tabcolsep}{2.2pt}
\caption{Precision (P), recall (R), and F1 (\%) of the passages that each method reads, measured against the gold evidence, with \haiku as the LLM. AS: Agentic Search; I-RAG: Interact-RAG. Best P, R, and F1 per dataset in \textbf{bold}.}
\label{tab:retrieval-haiku}
\resizebox{\columnwidth}{!}{%
\begin{tabular}{l >{\hspace{4pt}}ccc >{\hspace{4pt}}ccc >{\hspace{4pt}}ccc}
\toprule
 & \multicolumn{3}{c}{WixQA} & \multicolumn{3}{c}{QASPER} & \multicolumn{3}{c}{FinanceBench} \\
\cmidrule(lr){2-4}\cmidrule(lr){5-7}\cmidrule(lr){8-10}
Method & P & R & F1 & P & R & F1 & P & R & F1 \\
\midrule
\rowcolor{famVanilla}\multicolumn{10}{l}{\textit{Vanilla}} \\
\rowcolor{famVanilla}\hspace{0.8em}RAG ($k{=}5$) & 20.5 & 64.7 & 31.1 & 21.6 & 67.7 & 32.8 & 8.0 & 34.7 & 13.0 \\
\rowcolor{famAgentic}\multicolumn{10}{l}{\textit{Agentic}} \\
\rowcolor{famAgentic}\hspace{0.8em}AS & 15.1 & 70.8 & 24.9 & 17.3 & \textbf{85.5} & 28.8 & 11.5 & 71.0 & 19.8 \\
\rowcolor{famAgentic}\hspace{0.8em}I-RAG & 9.1 & 60.9 & 15.8 & 13.8 & 63.9 & 22.7 & 8.8 & 66.6 & 15.5 \\
\rowcolor{famAgentic}\hspace{0.8em}PageIndex & \textbf{46.2} & 52.5 & 49.1 & \textbf{54.9} & 50.4 & 52.6 & 33.1 & 56.8 & 41.8 \\
\addlinespace[1.5pt]
\rowcolor{famOurs}\ourmethod & 40.1 & \textbf{78.5} & \textbf{53.1} & 47.9 & 80.6 & \textbf{60.1} & \textbf{44.7} & \textbf{73.2} & \textbf{55.5} \\
\bottomrule
\end{tabular}%
}
\end{table}

\subsection{Cost Analysis}
\label{sec:cost}
% \ycom{App.: please add a section \texttt{\textbackslash label\{app:cost\}} with Figure \texttt{fig:latency} (figures/fig\_latency.tex), Table \texttt{tab:online-cost} (figures/table\_online\_cost.tex), and Table \texttt{tab:cost} (figures/table\_cost.tex).}

\noindent \textbf{Online Cost:}
We observe that on WixQA, QASPER, and FinanceBench, \ourmethod costs $5.7$--$13.3$ cents per question, more than Agentic Search ($1.3$--$6.3$ cents).
This is because the agent reads the string of the induced sub-forest that also includes the summaries and keywords.
On \entdocs, whose nodes carry only titles, this string is short, and \ourmethod costs about the same as Agentic Search ($6.0$ vs.\ $6.2$ cents).
In return, it improves accuracy by $2.4$--$10.5$ points over Agentic Search.
See fig.~\ref{fig:pareto} in App.~\ref{app:cost}. It plots accuracy against the mean API cost per question for the \haiku LLM and compares latency and cost for other baselines.
We observe that \ourmethod is the most accurate method across all datasets and lies on the Pareto frontier.

\noindent \textbf{Offline Cost:}
App.~\ref{app:cost} reports the one-time cost of preprocessing each corpus, at the API prices of \haiku (Table~\ref{tab:model-config}).
\ourmethod and PageIndex share the same preprocessing, \viz, inducing ToCs when needed and generating node metadata.
It costs \$$20.4$ on WixQA and \$$17.0$ on QASPER.
On FinanceBench, inducing the ToCs of $364$ long filings raises it to \$$773$.
On \entdocs, it costs nothing since we use the existing structure and only the node titles.
In contrast, the graph-based methods run LLM extraction over the entire corpus.
This costs \$$86$--$107$ on WixQA, and grows linearly with the size of the corpus, to an estimated \$$1.3$K--\$$1.7$K on FinanceBench and \$$106$K--\$$131$K on \entdocs.
This is why we do not run them on \entdocs.
% \ycom{In Table \texttt{tab:cost}, please fill in the QASPER cost of the graph-based methods (marked TBD) by scaling linearly from WixQA. Also check the \entdocs estimates: they are $357\times$ the WixQA cost, whereas \entdocs has about $1{,}233\times$ the tokens of WixQA, which would give about \$$106$K--\$$131$K.}

\section{Conclusion}
We presented \ourmethod, which combines content-based retrieval with structure-aware navigation.
Rather than showing the agent the retrieved chunks, or committing to a single document before reading any content, \ourmethod uses the retrieved chunks to extract query-specific sub-trees from the existing structure of the corpus, and lets the agent decide what to read.
On financial, scientific, and customer-support benchmarks, it achieves the highest accuracy against vanilla, graph-based, and agentic baselines, and balances the precision and recall of the text it reads.
On \entdocsbench, with $2.84$ million webpages, it improves accuracy by $6.6$--$11.4$ points over the strongest baseline, even though its nodes carry only their titles.
%and it needs no LLM preprocessing.
% \ourmethod costs more per question than Agentic Search, since the agent reads the string of the induced sub-trees after every search.
% Future work could reduce this cost by compacting the string representation, and study richer node metadata at the scale of millions of pages.

\section{Limitations}
\label{sec:limitations}
\ourmethod accumulates fetched nodes until it reaches its context budget and currently there is no support of evidence
compaction or eviction of the chat history. While \ourmethod preserves the hierarchical structure, it ignores any existing cross links between nodes at the same level in the tree. 
% \ourmethod also costs more and takes longer per query than single-shot RAG
% (Figure~\ref{fig:pareto} and Table~\ref{tab:cost}). Its advantage is therefore
% most useful when answer quality justifies iterative navigation. 
Finally, the method depends on a usable document hierarchy. Structure induction can recover one from documents such as SEC filings, but weak headings or poor parsing can
limit the quality of the map the agent sees.
\section*{Reproducibility Statement}
We include the information needed to independently reproduce our
experiments in the paper and appendix. The \ourmethod algorithm and
experimental protocol are described in
Sections~\ref{sec:ourmethod} and~\ref{sec:setup}, respectively.
Details of corpus processing, document-tree construction, and the
\entdocsbench dataset are provided in
Appendices~\ref{sec:corpus-extraction},
\ref{sec:stats}, and~\ref{sec:entdocs-creation}. Our implementation
choices, including node-aligned chunking, hybrid retrieval,
reciprocal-rank fusion, and the corpus-level PageIndex adaptation, are
documented in Appendices~\ref{app:rit-implementation}
and~\ref{app:pageindex}. We also report the model and decoding
settings, hyperparameter selection procedure, and cost assumptions in
Appendices~\ref{sec:model-config}, \ref{app:ablations},
and~\ref{app:cost}. The prompts used by the agents and evaluators are
included in Appendix~\ref{app:prompts}, and the answer-quality and
retrieval metrics are defined in
Appendices~\ref{app:quality} and~\ref{app:retrieval}.

% References (author-year, natbib) built from refs.bib via BibTeX.
% acl.sty sets \bibliographystyle{acl_natbib}
\bibliography{refs}

% Appendix: corpus extraction, statistics tables, prompts.
% ---------------------------------------------------------------------------
% \ourmethod — Appendix
%   A. Corpus-specific structure extraction
%   B. Dataset and structure statistics (Tables)
%   C. Prompts (dynamic-ToC agent + per-dataset judges)
% Requires in the preamble: \usepackage{listings}
% ---------------------------------------------------------------------------

\appendix

% ===========================================================================
\section{Corpus-Specific Structure Extraction}
\label{sec:corpus-extraction}

Offline, \ourmethod extracts the corpus structure by combining inter-document hierarchy (how
documents relate to one another) with intra-document structure induced inside
each document (how a document is subdivided into navigable sections).
The pipeline is corpus-agnostic except for the parsing step that extracts
nodes and parent links from each corpus's native form. The following describes the extraction pipelines for each corpus. Resulting statistics are in
Tables~\ref{tab:corpus-stats}--\ref{tab:structure-stats}.

\phead{WixQA}
Inter-document hierarchy is recovered from URL structure. Each article's title and URL identifies its product ($170$ distinct products). An LLM groups those
products into $16$ coarse families, giving a
three-level backbone family $\rightarrow$ product $\rightarrow$ article.
Intra-document structure is induced by splitting each article into sections and
subsections using its HTML heading tags.
Over the $6{,}221$ help articles this yields $31{,}733$ nodes; article/section body
nodes are readable, while the $16$ family and $170$ product nodes act as navigational
containers.

\phead{FinanceBench}
We treat the full set of $364$ SEC filings (10-Ks, 10-Qs, 8-Ks, and annual reports)
as a single corpus. Although the benchmark's $150$ questions draw on only $84$ of
these documents, we index all $364$ so the remaining filings act as distractors,
testing retrieval under realistic corpus scale rather than an oracle-filtered set.
The corpus has no authored inter-document hierarchy and hence we treat each filing as independent.
Intra-document structure must therefore be induced for every filing. Each
native PDF table of contents is too coarse for node-level navigation (individual
entries often span tens of pages), so we process each filing with the PageIndex
pipeline \footnote{\label{fn:pageindex}
\href{https://github.com/VectifyAI/PageIndex/tree/fb3f6e441a0741120da7e73b53ef9246d1642215}
{\nolinkurl{https://github.com/VectifyAI/PageIndex}}} to produce a finer-grained section hierarchy.

\phead{QASPER}
Each paper's section hierarchy is recovered directly from its structured markup.
Section and subsection headings become tree nodes, giving a shallow (depth $\le 4$)
per-paper hierarchy without additional induction. There is no inter-document hierarchy,
as papers are independent research artifacts.
Section body nodes are readable. Heading-only container nodes serve as navigational
anchors. Because the native section headings are already clean and informative, we
only generate node summaries and keywords but do not rewrite node
titles.

\phead{\entdocs}
The \entdocs corpus is derived from the IBM Documentation portal.\footnote{\url{https://www.ibm.com/docs/en/products}}
Starting from the portal homepage, we crawl the documentation while preserving the
existing product navigation trees.
The structure is entirely inter-document. Each product is organised as a
hierarchy of webpages connected through the sitemap.
There is no intra-document structure to induce, as each webpage corresponds to a
single node and is not further subdivided.
We combine the per-product trees under a virtual root to obtain a unified corpus
containing $2.84$M webpages, with a maximum depth of $17$.

\noindent To create the virtual document for PageIndex, we use greedy DFS packing on the global ToC to fit as many webpages into the virtual doc $\doc{v}$ till its corresponding string representation $\strrep(\toc{v})$ reaches a limit of 100k tokens. This is done to ensure that the $\strrep(\toc{v})$ fits inside the context window of the evaluated models, with enough room for reading calls.

\section{Retrieval Implementation}
\subsection{\ourmethod}
\label{app:rit-implementation}
We chunk each node of the corpus $\corpus$ into chunks of fixed token size of 512. \ourmethod uses hybrid retriever for scoring a chunk's similarity to the query. To return the top-$k$ chunks, it retrieves the top $k'$ chunks from BM25 and from the dense
index, fuses the two rankings with reciprocal rank fusion \citep{cormack2009rrf} with fusion constant $\kappa=60$, the standard RRF setting combining both the lexical overlap and semantic signals.

% ===========================================================================
\subsection{PageIndex}
\label{app:pageindex}

We implement the corpus-level PageIndex following the methodology described in ~\citep{pageindex}. We first retrieve the top $k$ chunks with vector search. For a filing
$D$ with $n_D$ retrieved chunks, we compute
$\operatorname{DocScore}(D) = \frac{1}{\sqrt{n_D+1}}
\sum_{c}\operatorname{ChunkScore}(c)$, summing over the retrieved chunks $c$ that belong to $D$. We select the highest-scoring
filing, retrieve its table of contents, and pass
that structure to the PageIndex agent for within-document navigation.

% ===========================================================================
\section{Dataset and Structure Statistics}
\label{sec:stats}

Table~\ref{tab:corpus-stats} reports corpus and benchmark sizes.
Table~\ref{tab:structure-stats} reports the extracted corpus structure.

\begin{table*}[t]
\centering
\small
\caption{Corpus and evaluation-set statistics. \textbf{Docs} is the
number of source pdfs or webpages indexed; \textbf{Chunks} is the number of
retrieval chunks after node-aligned chunking;
\textbf{Q} is the number of questions evaluated.}
\label{tab:corpus-stats}
\begin{tabular}{llrrrr}
\toprule
Corpus & Node $=$ & Docs & Chunks & Tokens & Q \\
\midrule
WixQA        & article section \  & 6{,}221       & 32{,}032      & 3.09\,M & 200        \\
FinanceBench & PDF page span  & 364           & 129{,}973     & 48.5\,M & 150        \\
\entdocs        & doc.\ webpage  & 2{,}839{,}892 & 8{,}993{,}388 & 3.81\,B & 317 \\
QASPER       & paper section  & 1{,}585       & 31{,}014      & 10.6\,M & 1372  \\
\bottomrule
\end{tabular}
\end{table*}

\begin{table*}[t]
\centering
\small
\caption{Hierarchical document structure per corpus. Depth is from the document root; node
size is a node's own-text token count.}
\label{tab:structure-stats}
\begin{tabular}{lrrrrrr}
\toprule
Corpus & Total & Max & \multicolumn{3}{c}{Node size (tok)} \\
       & nodes & depth & mean & med. & p95 \\
\midrule
WixQA        & 31{,}733      & 5  & 84  & 60  & 219     \\
FinanceBench & 79{,}658     & 8  & 602 & 274 & 2{,}368 \\
\entdocs       & 2{,}934{,}928 & 17 & 1{,}020 & 182 & 1{,}960 \\
QASPER       & 24{,}350      & 4  & 362 & 262 & 977     \\
\bottomrule
\end{tabular}
\end{table*}

% ===========================================================================
\section{Model and Inference Configuration}
\label{sec:model-config}

Table~\ref{tab:model-config} lists the language models used for generation and
offline preprocessing, together with decoding settings and the per-token prices
used for cost accounting in Table~\ref{tab:cost}.

\begin{table*}[t]
\centering
\small
\caption{Models and decoding settings used in the experiments.
\textbf{Role} identifies how each model is used in the pipeline.
\textbf{Temperature} and \textbf{Max tokens} report the generation
settings. \textbf{Input/Output price} gives the effective API cost in
USD per one million tokens.}
\label{tab:model-config}
% tabularx: the long ``Key decoding settings'' column (X, elastic, ragged-right)
% wraps to fit the single-column text width, keeping the body font consistent
% with the rest of the paper instead of shrinking it via \resizebox.
\begin{tabularx}{\textwidth}{ll >{\raggedright\arraybackslash}X r}
\toprule
Model & Role & Key decoding settings & \$/1M (in\,/\,out) \\
\midrule
\haikofull & Generation + preprocessing & default sampling ($512$ output tok.\ for prep) & $1.00$\,/\,$5.00$ \\
\gptmini       & Generation                 & temperature $1$; medium reasoning effort & $0.25$\,/\,$2.00$ \\
\lunafull (\luna)   & Generation                 & temperature $1$; non-reasoning & $1.00$\,/\,$6.00$ \\
\qwen        & Preprocessing              & non-thinking; $T{=}0.7$, top-$p{=}0.8$, top-$k{=}20$, min-$p{=}0$; YaRN $64$k & $0.08$\,/\,$0.28$ \\
\judgemodel & Evaluation (binary + G-E judges) & $T{=}0$ & -- \\
\bottomrule
\end{tabularx}
\end{table*}

\section{Cost and Latency Analysis}
\label{app:cost}
% Generated TikZ bodies (analysis/gen_pareto_figure.py, gen_latency_figure.py) carry no float
% or caption, so the figure* wrappers live here and survive regeneration.
\begin{figure*}[t]
\centering
\input{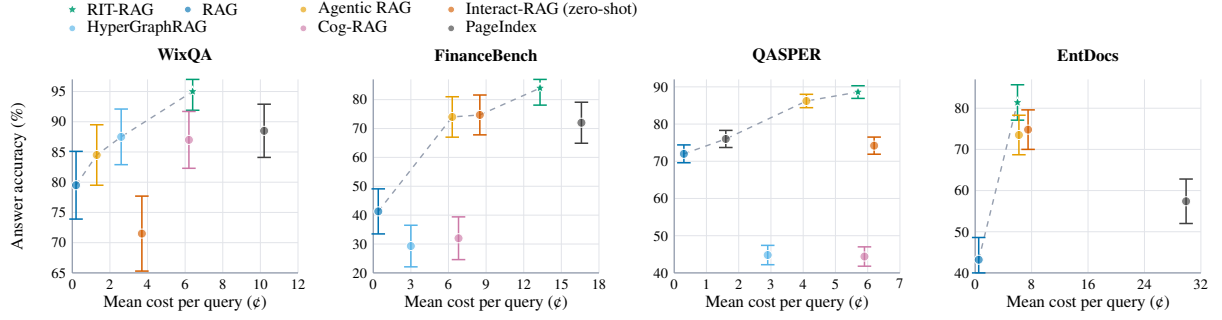}
\caption[Accuracy vs.\ per-query cost]{Answer accuracy vs.\ mean API cost per question (US cents, \haiku generation); bars are 95\% Wilson intervals. The graph baselines are not run on \entdocs (Table~\ref{tab:cost}).}
\label{fig:pareto}
\end{figure*}
\begin{figure*}[t]
\centering
\input{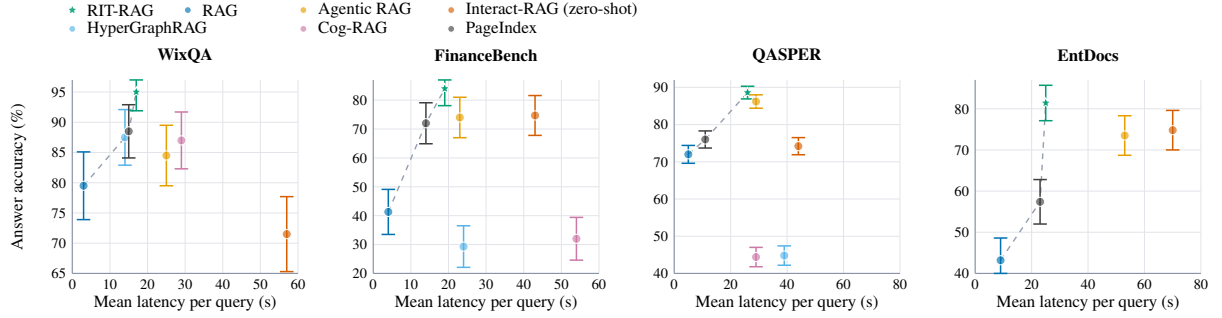}
\caption[Accuracy vs.\ per-query latency]{Answer accuracy vs.\ mean latency per question (seconds, \haiku generation); bars are 95\% Wilson intervals.}
\label{fig:latency}
\end{figure*}
\begin{table*}[t]
\centering\small
\setlength{\tabcolsep}{5pt}
\caption{Per-query online trade-offs (Haiku generation). Acc.: binary-judge accuracy (\%); \textcent/q: mean API cost (US cents); s/q: mean latency (s). The graph baselines are not run on EntDocs (prohibitive preprocessing, Table~\ref{tab:cost}).}
\label{tab:online-cost}
\adjustbox{max width=\textwidth}{%
\begin{tabular}{l ccc ccc ccc ccc}
\toprule
 & \multicolumn{3}{c}{WixQA ($n{=}200$)} & \multicolumn{3}{c}{QASPER ($n{=}1372$)} & \multicolumn{3}{c}{FinanceBench ($n{=}150$)} & \multicolumn{3}{c}{EntDocs ($n{=}317$)} \\
\cmidrule(lr){2-4}\cmidrule(lr){5-7}\cmidrule(lr){8-10}\cmidrule(lr){11-13}
Method & Acc. & \textcent/q & s/q & Acc. & \textcent/q & s/q & Acc. & \textcent/q & s/q & Acc. & \textcent/q & s/q \\
\midrule
RAG ($k{=}5$) & 79.5 & 0.2 & 3 & 72.0 & 0.3 & 5 & 41.3 & 0.4 & 4 & 43.2 & 0.5 & 9 \\
HyperGraphRAG & 87.5 & 2.6 & 14 & 44.8 & 2.9 & 39 & 29.3 & 3.0 & 24 & -- & -- & -- \\
Cog-RAG & 87.0 & 6.2 & 29 & 44.4 & 5.9 & 29 & 32.0 & 6.8 & 54 & -- & -- & -- \\
Agentic Search & 84.5 & 1.3 & 25 & 86.2 & 4.1 & 29 & 74.0 & 6.3 & 23 & 73.5 & 6.2 & 53 \\
Interact-RAG & 71.5 & 3.7 & 57 & 74.2 & 6.2 & 44 & 74.7 & 8.5 & 43 & 74.8 & 7.5 & 70 \\
PageIndex & 93.0 & 9.9 & 14 & 76.0 & 1.6 & 11 & 72.0 & 16.6 & 14 & 57.4 & 29.9 & 23 \\
\ourmethod & \textbf{95.0} & 6.4 & 17 & \textbf{88.6} & 5.7 & 26 & \textbf{84.0} & 13.3 & 19 & \textbf{81.4} & 6.0 & 25 \\
\bottomrule
\end{tabular}%
}
\end{table*}

% Appendix table (26 Sep, goes to app:cost). Methods without preprocessing (vanilla RAG, Agentic Search, Interact-RAG) cost nothing and are omitted.
% RIT-RAG / PageIndex rows re-priced 2026-10-13 at Haiku $1/$5 per 1M tokens from recorded token counts:
% WixQA analysis/wixqa/2026-09-15_matrix.xlsx Preprocessing; QASPER benchmarks/qasper/data/parsed/summaries_cost.json;
% FinanceBench analysis/finance_bench/2026-09-17_matrix.xlsx Preprocessing (LLM tokens at $1/$5 + embeddings $0.97).
% Full version: _review/tables_before_26sep/table_cost.tex
% Graph rows: daggered cells = WixQA cost x corpus-token ratio from Table 5 (QASPER 10.6M/3.09M = 3.43,
% FinanceBench 48.5M/3.09M = 15.70, EntDocs 3.81B/3.09M = 1233). Fixed 2026-10-13: EntDocs had used an untraceable x357.
\begin{table}[t]
\centering\small
\setlength{\tabcolsep}{3pt}
\caption{One-time preprocessing cost (USD) at the prices of \haiku (Table~\ref{tab:model-config}). The other baselines need no preprocessing. $^\dagger$Estimated by scaling linearly from WixQA.}
\label{tab:cost}
\adjustbox{max width=\linewidth}{%
\begin{tabular}{l rrrr}
\toprule
Method & WixQA & QASPER & FinanceBench & \entdocs \\
\midrule
HyperGraphRAG & \$85.9  & ${\sim}$\$294.70$^\dagger$ & ${\sim}$\$1.3K$^\dagger$ & ${\sim}$\$105.9K$^\dagger$ \\
Cog-RAG       & \$106.6 & ${\sim}$\$365.70$^\dagger$ & ${\sim}$\$1.7K$^\dagger$ & ${\sim}$\$131.4K$^\dagger$ \\
PageIndex     & \$20.4  & \$17.0 & \$773.1 & \$0 \\
\ourmethod    & \$20.4  & \$17.0 & \$773.1 & \$0 \\
\bottomrule
\end{tabular}%
}
\end{table}

Table~\ref{tab:cost} reports the one-time preprocessing cost for each method at
the API prices listed in Table~\ref{tab:model-config}.

At query time, \ourmethod costs more than vanilla RAG (0.2–0.5 cents per question), but improves accuracy by 15.5–
42.7 points. \ourmethod is also faster than Agentic Search (17–26 vs. 23–53 seconds) and Interact-
RAG (43–70 seconds) (Figure ~\ref{fig:latency}).
%This is because \ourmethod makes only about three tool calls per question, whereas Agentic Search and Interact-RAG make 3.7 and 3.6 tool calls on average, respectively.
PageIndex
is the fastest agentic method (11–23 seconds), and on QASPER, where each document is small, it
is also the cheapest (1.6 cents). However, it trails RIT-RAG in accuracy on all four datasets, as it
cannot change the document once it commits to it.
On \entdocs, PageIndex is the most expensive method ($29.9$ cents per question) as the table of contents of the virtual document it selects spans thousands of pages on average.
% so it reads about $289$K input tokens per question, compared with $53$K for \ourmethod.

\section{Answer-Quality Metrics}
\label{app:quality}

We report lexical token recall and all seven G-E dimensions~\citep{luo2025hypergraphrag}
(correctness, relevance, factuality, comprehensiveness, knowledgeability, logical
coherence, and diversity; each normalized to $0$--$1$) for every dataset, using one
preprocessing model per corpus and all three generation models (\haiku, \gptmini,
and \luna; Tables~\ref{tab:quality-wixqa}--\ref{tab:quality-qasper}).
All G-E dimensions across every method and generation model are scored by the same
(\judgemodel) judge so that the values are comparable within and across rows.

\begin{table*}[t]
\centering\small
\setlength{\tabcolsep}{4pt}
\caption{WixQA answer quality by method and generation model. G-E dimensions (each normalized $0$--$1$): Corr(ectness), Rel(evance), Fact(uality), Compr(ehensiveness), Know(ledgeability), Coher(ence, logical), Div(ersity). Recall is lexical token recall. Best per column in \textbf{bold}.}
\label{tab:quality-wixqa}
\begin{tabular}{llcccccccc}
\toprule
Method & Gen & Recall & Corr & Rel & Fact & Compr & Know & Coher & Div \\
\midrule
\rowcolor{famVanilla} RAG ($k{=}5$) & Haiku & 0.427 & 0.740 & 0.821 & 0.779 & 0.658 & 0.745 & 0.948 & 0.491 \\
\rowcolor{famVanilla}  & GPT-5-mini & 0.411 & 0.792 & 0.883 & 0.796 & 0.739 & 0.851 & 0.929 & 0.652 \\
\rowcolor{famVanilla}  & Luna & 0.256 & 0.778 & 0.850 & 0.831 & 0.632 & 0.742 & 0.907 & 0.528 \\
\addlinespace[1.5pt]
\rowcolor{famGraph} HyperGraphRAG & Haiku & 0.632 & 0.858 & 0.926 & 0.860 & 0.823 & 0.889 & 0.945 & 0.781 \\
\rowcolor{famGraph}  & GPT-5-mini & \textbf{0.655} & 0.969 & \textbf{0.989} & 0.975 & 0.959 & 0.978 & 0.997 & 0.932 \\
\rowcolor{famGraph}  & Luna & 0.514 & 0.919 & 0.974 & 0.935 & 0.882 & 0.922 & 0.976 & 0.792 \\
\addlinespace[1.5pt]
\rowcolor{famGraph} Cog-RAG & Haiku & 0.614 & 0.849 & 0.919 & 0.879 & 0.829 & 0.892 & 0.939 & 0.834 \\
\rowcolor{famGraph}  & GPT-5-mini & 0.647 & \textbf{0.980} & \textbf{0.989} & \textbf{0.980} & \textbf{0.966} & \textbf{0.992} & \textbf{0.999} & \textbf{0.953} \\
\rowcolor{famGraph}  & Luna & 0.525 & 0.946 & 0.984 & 0.953 & 0.907 & 0.960 & 0.987 & 0.831 \\
\addlinespace[1.5pt]
\rowcolor{famAgentic} Agentic Search & Haiku & 0.581 & 0.814 & 0.926 & 0.828 & 0.777 & 0.862 & 0.942 & 0.660 \\
\rowcolor{famAgentic}  & GPT-5-mini & 0.527 & 0.911 & 0.952 & 0.921 & 0.877 & 0.929 & 0.978 & 0.727 \\
\rowcolor{famAgentic}  & Luna & 0.404 & 0.922 & 0.968 & 0.938 & 0.840 & 0.921 & 0.990 & 0.679 \\
\addlinespace[1.5pt]
\rowcolor{famAgentic} Interact-RAG & Haiku & 0.391 & 0.731 & 0.912 & 0.798 & 0.766 & 0.855 & 0.845 & 0.726 \\
\rowcolor{famAgentic}  & GPT-5-mini & 0.207 & 0.849 & 0.894 & 0.892 & 0.861 & 0.912 & 0.896 & 0.906 \\
\rowcolor{famAgentic}  & Luna & 0.352 & 0.876 & 0.928 & 0.914 & 0.865 & 0.895 & 0.954 & 0.811 \\
\addlinespace[1.5pt]
\rowcolor{famAgentic} PageIndex & Haiku & 0.619 & 0.851 & 0.917 & 0.877 & 0.829 & 0.884 & 0.945 & 0.715 \\
\rowcolor{famAgentic}  & GPT-5-mini & 0.548 & 0.911 & 0.945 & 0.923 & 0.878 & 0.917 & 0.982 & 0.772 \\
\rowcolor{famAgentic}  & Luna & 0.462 & 0.925 & 0.952 & 0.934 & 0.867 & 0.918 & 0.969 & 0.714 \\
\addlinespace[1.5pt]
\rowcolor{famOurs} \ourmethod & Haiku & 0.638 & 0.872 & 0.958 & 0.876 & 0.869 & 0.923 & 0.958 & 0.650 \\
\rowcolor{famOurs}  & GPT-5-mini & 0.587 & 0.948 & 0.980 & 0.965 & 0.930 & 0.969 & 0.994 & 0.677 \\
\rowcolor{famOurs}  & Luna & 0.474 & 0.941 & 0.980 & 0.956 & 0.909 & 0.953 & 0.986 & 0.760 \\
\bottomrule
\end{tabular}
\end{table*}

\begin{table*}[t]
\centering\small
\setlength{\tabcolsep}{4pt}
\caption{FinanceBench answer quality by method and generation model. G-E dimensions (each normalized $0$--$1$): Corr(ectness), Rel(evance), Fact(uality), Compr(ehensiveness), Know(ledgeability), Coher(ence, logical), Div(ersity). Recall is lexical token recall. Best per column in \textbf{bold}.}
\label{tab:quality-finance}
\begin{tabular}{llcccccccc}
\toprule
Method & Gen & Recall & Corr & Rel & Fact & Compr & Know & Coher & Div \\
\midrule
\rowcolor{famVanilla} RAG ($k{=}5$) & Haiku & 0.370 & 0.791 & 0.933 & 0.771 & 0.881 & 0.835 & 0.988 & 0.433 \\
\rowcolor{famVanilla}  & GPT-5-mini & 0.278 & 0.705 & 0.910 & 0.733 & 0.843 & 0.837 & 0.985 & 0.455 \\
\rowcolor{famVanilla}  & Luna & 0.255 & 0.667 & 0.918 & 0.677 & 0.803 & 0.760 & 0.993 & 0.407 \\
\addlinespace[1.5pt]
\rowcolor{famGraph} HyperGraphRAG & Haiku & 0.386 & 0.515 & 0.807 & 0.546 & 0.682 & 0.687 & 0.959 & 0.599 \\
\rowcolor{famGraph}  & GPT-5-mini & 0.296 & 0.574 & 0.828 & 0.588 & 0.645 & 0.648 & 0.737 & 0.407 \\
\rowcolor{famGraph}  & Luna & 0.338 & 0.693 & 0.935 & 0.683 & 0.697 & 0.737 & 0.778 & 0.401 \\
\addlinespace[1.5pt]
\rowcolor{famGraph} Cog-RAG & Haiku & 0.417 & 0.551 & 0.851 & 0.582 & 0.721 & 0.763 & 0.985 & 0.581 \\
\rowcolor{famGraph}  & GPT-5-mini & 0.336 & 0.650 & 0.835 & 0.632 & 0.682 & 0.745 & 0.809 & 0.462 \\
\rowcolor{famGraph}  & Luna & 0.386 & 0.691 & 0.899 & 0.728 & 0.695 & 0.754 & 0.753 & 0.431 \\
\addlinespace[1.5pt]
\rowcolor{famAgentic} Agentic Search & Haiku & 0.545 & 0.837 & 0.923 & 0.856 & 0.893 & 0.915 & 0.982 & 0.550 \\
\rowcolor{famAgentic}  & GPT-5-mini & 0.425 & 0.888 & 0.965 & 0.885 & 0.907 & 0.920 & 0.992 & 0.451 \\
\rowcolor{famAgentic}  & Luna & 0.422 & 0.863 & 0.941 & 0.865 & 0.895 & 0.914 & 0.988 & 0.465 \\
\addlinespace[1.5pt]
\rowcolor{famAgentic} Interact-RAG & Haiku & 0.444 & 0.821 & 0.927 & 0.817 & 0.741 & 0.831 & 0.655 & 0.371 \\
\rowcolor{famAgentic}  & GPT-5-mini & 0.300 & 0.799 & 0.915 & 0.808 & 0.683 & 0.799 & 0.636 & 0.301 \\
\rowcolor{famAgentic}  & Luna & 0.430 & 0.881 & 0.961 & 0.889 & 0.851 & 0.895 & 0.845 & 0.405 \\
\addlinespace[1.5pt]
\rowcolor{famAgentic} PageIndex & Haiku & 0.548 & 0.866 & 0.939 & 0.865 & 0.907 & 0.927 & 0.976 & \textbf{0.623} \\
\rowcolor{famAgentic}  & GPT-5-mini & 0.458 & 0.885 & 0.964 & 0.900 & 0.926 & 0.933 & 0.985 & 0.492 \\
\rowcolor{famAgentic}  & Luna & 0.300 & 0.869 & 0.971 & 0.873 & 0.907 & 0.920 & 0.992 & 0.519 \\
\addlinespace[1.5pt]
\rowcolor{famOurs} \ourmethod & Haiku & \textbf{0.579} & 0.913 & 0.949 & 0.908 & 0.935 & 0.947 & 0.980 & 0.620 \\
\rowcolor{famOurs}  & GPT-5-mini & 0.458 & 0.923 & 0.973 & 0.943 & \textbf{0.949} & \textbf{0.964} & 0.995 & 0.534 \\
\rowcolor{famOurs}  & Luna & 0.516 & \textbf{0.941} & \textbf{0.977} & \textbf{0.951} & 0.938 & 0.963 & \textbf{0.997} & 0.537 \\
\bottomrule
\end{tabular}
\end{table*}

\begin{table*}[t]
\centering\small
\setlength{\tabcolsep}{4pt}
\caption{\entdocsbench answer quality by method and generation model. G-E dimensions (each normalized $0$--$1$): Corr(ectness), Rel(evance), Fact(uality), Compr(ehensiveness), Know(ledgeability), Coher(ence, logical), Div(ersity). Recall is lexical token recall. Best per column in \textbf{bold}.}
\label{tab:quality-ibmz}
\begin{tabular}{llcccccccc}
\toprule
Method & Gen & Recall & Corr & Rel & Fact & Compr & Know & Coher & Div \\
\midrule
\rowcolor{famVanilla} RAG ($k{=}5$) & Haiku & 0.392 & 0.601 & 0.698 & 0.626 & 0.607 & 0.557 & 0.928 & 0.379 \\
\rowcolor{famVanilla}  & GPT-5-mini & 0.288 & 0.709 & 0.801 & 0.721 & 0.655 & 0.697 & 0.880 & 0.457 \\
\rowcolor{famVanilla}  & Luna & 0.218 & 0.655 & 0.734 & 0.664 & 0.564 & 0.608 & 0.827 & 0.385 \\
\addlinespace[1.5pt]
\rowcolor{famAgentic} Agentic Search & Haiku & 0.559 & 0.827 & 0.868 & 0.832 & 0.777 & 0.839 & 0.918 & 0.470 \\
\rowcolor{famAgentic}  & GPT-5-mini & 0.369 & 0.834 & 0.883 & 0.851 & 0.759 & 0.836 & 0.929 & 0.502 \\
\rowcolor{famAgentic}  & Luna & 0.374 & 0.852 & 0.891 & 0.863 & 0.790 & 0.845 & 0.924 & 0.548 \\
\addlinespace[1.5pt]
\rowcolor{famAgentic} PageIndex & Haiku & 0.538 & 0.696 & 0.790 & 0.750 & 0.673 & 0.731 & 0.891 & 0.546 \\
\rowcolor{famAgentic}  & GPT-5-mini & 0.373 & 0.722 & 0.811 & 0.767 & 0.657 & 0.701 & 0.942 & 0.476 \\
\rowcolor{famAgentic}  & Luna & 0.373 & 0.785 & 0.869 & 0.806 & 0.728 & 0.757 & 0.959 & 0.541 \\
\addlinespace[1.5pt]
\rowcolor{famOurs} \ourmethod & Haiku & \textbf{0.602} & 0.891 & 0.924 & 0.890 & 0.860 & 0.892 & 0.933 & 0.551 \\
\rowcolor{famOurs}  & GPT-5-mini & 0.447 & 0.910 & 0.934 & 0.914 & 0.854 & 0.910 & 0.968 & 0.522 \\
\rowcolor{famOurs}  & Luna & 0.443 & \textbf{0.937} & \textbf{0.966} & \textbf{0.938} & \textbf{0.898} & \textbf{0.942} & \textbf{0.971} & \textbf{0.637} \\
\bottomrule
\end{tabular}
\end{table*}

\begin{table*}[t]
\centering\small
\setlength{\tabcolsep}{4pt}
\caption{QASPER answer quality by method and generation model. G-E dimensions (each normalized $0$--$1$): Corr(ectness), Rel(evance), Fact(uality), Compr(ehensiveness), Know(ledgeability), Coher(ence, logical), Div(ersity). Recall is lexical token recall. Best per column in \textbf{bold}.}
\label{tab:quality-qasper}
\begin{tabular}{llcccccccc}
\toprule
Method & Gen & Recall & Corr & Rel & Fact & Compr & Know & Coher & Div \\
\midrule
\rowcolor{famVanilla} RAG ($k{=}5$) & Haiku & 0.680 & 0.836 & 0.925 & 0.823 & 0.869 & 0.822 & 0.995 & 0.537 \\
\rowcolor{famVanilla}  & GPT-5-mini & 0.611 & 0.833 & 0.913 & 0.828 & 0.835 & 0.846 & 0.987 & 0.574 \\
\rowcolor{famVanilla}  & Luna & 0.569 & 0.789 & 0.885 & 0.792 & 0.779 & 0.795 & 0.978 & 0.516 \\
\addlinespace[1.5pt]
\rowcolor{famGraph} HyperGraphRAG & Haiku & 0.612 & 0.651 & 0.765 & 0.653 & 0.650 & 0.694 & 0.944 & 0.646 \\
\rowcolor{famGraph}  & GPT-5-mini & 0.530 & 0.729 & 0.841 & 0.734 & 0.693 & 0.760 & 0.939 & 0.603 \\
\rowcolor{famGraph}  & Luna & 0.542 & 0.752 & 0.838 & 0.756 & 0.707 & 0.776 & 0.934 & 0.605 \\
\addlinespace[1.5pt]
\rowcolor{famGraph} Cog-RAG & Haiku & 0.557 & 0.588 & 0.724 & 0.604 & 0.600 & 0.636 & 0.920 & 0.606 \\
\rowcolor{famGraph}  & GPT-5-mini & 0.534 & 0.745 & 0.848 & 0.746 & 0.709 & 0.770 & 0.950 & 0.615 \\
\rowcolor{famGraph}  & Luna & 0.525 & 0.747 & 0.843 & 0.755 & 0.719 & 0.782 & 0.949 & 0.633 \\
\addlinespace[1.5pt]
\rowcolor{famAgentic} Agentic Search & Haiku & 0.822 & 0.897 & 0.941 & 0.892 & 0.907 & 0.935 & 0.989 & 0.685 \\
\rowcolor{famAgentic}  & GPT-5-mini & 0.714 & 0.897 & 0.948 & 0.900 & 0.900 & 0.923 & \textbf{0.997} & 0.578 \\
\rowcolor{famAgentic}  & Luna & 0.696 & 0.896 & 0.947 & 0.903 & 0.890 & 0.922 & 0.990 & 0.657 \\
\addlinespace[1.5pt]
\rowcolor{famAgentic} Interact-RAG & Haiku & 0.672 & 0.798 & 0.888 & 0.792 & 0.764 & 0.809 & 0.946 & 0.560 \\
\rowcolor{famAgentic}  & GPT-5-mini & 0.499 & 0.773 & 0.873 & 0.775 & 0.748 & 0.801 & 0.898 & 0.568 \\
\rowcolor{famAgentic}  & Luna & 0.555 & 0.830 & 0.911 & 0.829 & 0.802 & 0.848 & 0.951 & 0.619 \\
\addlinespace[1.5pt]
\rowcolor{famAgentic} PageIndex & Haiku & 0.736 & 0.868 & 0.902 & 0.876 & 0.860 & 0.880 & 0.986 & 0.696 \\
\rowcolor{famAgentic}  & GPT-5-mini & 0.644 & 0.848 & 0.927 & 0.862 & 0.847 & 0.859 & 0.988 & 0.621 \\
\rowcolor{famAgentic}  & Luna & 0.641 & 0.862 & 0.916 & 0.860 & 0.848 & 0.875 & 0.981 & 0.685 \\
\addlinespace[1.5pt]
\rowcolor{famOurs} \ourmethod & Haiku & \textbf{0.836} & 0.915 & 0.945 & 0.908 & 0.924 & \textbf{0.951} & 0.990 & \textbf{0.755} \\
\rowcolor{famOurs}  & GPT-5-mini & 0.737 & \textbf{0.929} & \textbf{0.963} & 0.929 & \textbf{0.935} & 0.949 & \textbf{0.997} & 0.613 \\
\rowcolor{famOurs}  & Luna & 0.715 & 0.923 & 0.961 & \textbf{0.930} & 0.923 & 0.947 & 0.993 & 0.726 \\
\bottomrule
\end{tabular}
\end{table*}

% ===========================================================================

\section{Retrieval Metrics}
\label{app:retrieval}

% Retrieval quality with \gptmini as the LLM. Numbers from figures/table_retrieval_{wixqa,qasper,finance}.tex
% (GPT-5-mini rows), times 100. F1 = 2PR/(P+R), computed from the rounded P and R.
% Graph-based methods are excluded on purpose: they also read the entity / hyperedge descriptions built in preprocessing.
% GENERATED by analysis/gen_retrieval_f1_tables.py — do not edit by hand.
\begin{table*}[t]
\centering\small
\setlength{\tabcolsep}{3.5pt}
\caption{Precision (P), recall (R), and F1 (\%) of the passages that each method reads, measured against the gold evidence, with \gptmini as the LLM. Best P, R, and F1 per dataset in \textbf{bold}.}
\label{tab:retrieval-gpt5mini}
\adjustbox{max width=\textwidth}{%
\begin{tabular}{ll @{\hspace{10pt}} ccc @{\hspace{10pt}} ccc @{\hspace{10pt}} ccc}
\toprule
\multicolumn{2}{@{}l}{\textbf{Dataset}} & \multicolumn{3}{c}{WixQA} & \multicolumn{3}{c}{QASPER} & \multicolumn{3}{c}{FinanceBench} \\
\cmidrule(lr){3-5}\cmidrule(lr){6-8}\cmidrule(lr){9-11}
Family & Method & P & R & F1 & P & R & F1 & P & R & F1 \\
\midrule
\addlinespace[1.5pt]
\rowcolor{famVanilla}\textit{Vanilla} & RAG ($k{=}5$) & 20.5 & 64.7 & 31.1 & 21.6 & 67.7 & 32.8 & 8.0 & 34.7 & 13.0 \\
\addlinespace[1.5pt]
\rowcolor{famAgentic}\textit{Agentic} & Agentic Search & 17.1 & 76.9 & 28.0 & 19.0 & 85.0 & 31.1 & 12.9 & 68.6 & 21.7 \\
\rowcolor{famAgentic} & Interact-RAG & 7.2 & 75.4 & 13.1 & 9.6 & 61.1 & 16.6 & 6.9 & 63.2 & 12.4 \\
\rowcolor{famAgentic} & PageIndex & \textbf{47.0} & 53.0 & \textbf{49.8} & \textbf{50.4} & 49.7 & \textbf{50.0} & \textbf{34.5} & 58.0 & \textbf{43.3} \\
\addlinespace[1.5pt]
\rowcolor{famOurs}\textit{Ours} & \ourmethod & 31.8 & \textbf{80.0} & 45.5 & 32.3 & \textbf{86.1} & 47.0 & 29.6 & \textbf{78.1} & 42.9 \\
\bottomrule
\end{tabular}%
}
\end{table*}

% Retrieval quality with \luna as the LLM. Numbers from figures/table_retrieval_{wixqa,qasper,finance}.tex
% (Luna rows), times 100. F1 = 2PR/(P+R), computed from the rounded P and R.
% Graph-based methods are excluded on purpose: they also read the entity / hyperedge descriptions built in preprocessing.
% GENERATED by analysis/gen_retrieval_f1_tables.py — do not edit by hand.
\begin{table*}[t]
\centering\small
\setlength{\tabcolsep}{3.5pt}
\caption{Precision (P), recall (R), and F1 (\%) of the passages that each method reads, measured against the gold evidence, with \luna as the LLM. Best P, R, and F1 per dataset in \textbf{bold}.}
\label{tab:retrieval-luna}
\adjustbox{max width=\textwidth}{%
\begin{tabular}{ll @{\hspace{10pt}} ccc @{\hspace{10pt}} ccc @{\hspace{10pt}} ccc}
\toprule
\multicolumn{2}{@{}l}{\textbf{Dataset}} & \multicolumn{3}{c}{WixQA} & \multicolumn{3}{c}{QASPER} & \multicolumn{3}{c}{FinanceBench} \\
\cmidrule(lr){3-5}\cmidrule(lr){6-8}\cmidrule(lr){9-11}
Family & Method & P & R & F1 & P & R & F1 & P & R & F1 \\
\midrule
\addlinespace[1.5pt]
\rowcolor{famVanilla}\textit{Vanilla} & RAG ($k{=}5$) & 20.5 & 64.7 & 31.1 & 21.6 & 67.7 & 32.8 & 8.0 & 34.7 & 13.0 \\
\addlinespace[1.5pt]
\rowcolor{famAgentic}\textit{Agentic} & Agentic Search & 16.6 & 75.4 & 27.2 & 19.8 & 85.6 & 32.2 & 13.1 & 72.8 & 22.2 \\
\rowcolor{famAgentic} & Interact-RAG & 12.6 & 74.8 & 21.6 & 13.4 & 68.1 & 22.4 & 10.3 & 67.3 & 17.9 \\
\rowcolor{famAgentic} & PageIndex & \textbf{46.4} & 47.7 & \textbf{47.0} & \textbf{49.3} & 43.9 & 46.4 & \textbf{31.6} & 50.4 & 38.8 \\
\addlinespace[1.5pt]
\rowcolor{famOurs}\textit{Ours} & \ourmethod & 32.4 & \textbf{80.1} & 46.1 & 34.4 & \textbf{86.4} & \textbf{49.2} & 30.1 & \textbf{77.1} & \textbf{43.3} \\
\bottomrule
\end{tabular}%
}
\end{table*}

We compute F1 as $2PR/(P+R)$, where $P$ and $R$ are the precision and recall of the passages that a method reads, measured against the gold evidence.
% We exclude the graph-based methods from the analysis in Section~\ref{sec:analysis}.
% Besides the raw chunks, they read the descriptions of the entities and hyperedges that they create during preprocessing, and their answer may come from these descriptions.
% Thus, measuring their precision and recall on the raw chunks alone would be unfair to them, and we cannot measure it on the descriptions.
With \haiku as the LLM (Table~\ref{tab:retrieval-haiku}), Agentic Search has the highest recall on QASPER ($85.5\%$), but a precision of only $11.5$--$17.3\%$, lower than even vanilla RAG on WixQA and QASPER.
PageIndex has the highest precision on WixQA ($46.2\%$) and QASPER ($54.9\%$), but its recall drops to $52.5\%$ on WixQA, $50.4\%$ on QASPER, and $56.8\%$ on FinanceBench.
On QASPER, \ourmethod is second on both measures, but close to the best: $80.6\%$ vs.\ $85.5\%$ recall, and $47.9\%$ vs.\ $54.9\%$ precision.
With \luna (Table~\ref{tab:retrieval-luna}), \ourmethod also has the highest F1 on QASPER, and FinanceBench, and is comparable with PageIndex ($46.1$ vs $47.0$).
With \gptmini (Table~\ref{tab:retrieval-gpt5mini}), PageIndex has a slightly higher F1 than \ourmethod on WixQA ($49.8\%$ vs.\ $45.5\%$), QASPER ($50.0\%$ vs.\ $47.0\%$) and FinanceBench ($43.3\%$ vs.\ $42.9\%$), although \ourmethod still has the highest recall on these three datasets.

\section{Complementarity Analysis}
\label{app:correct-overlap}

\begin{figure*}[t]
\centering
\includegraphics[width=\textwidth]{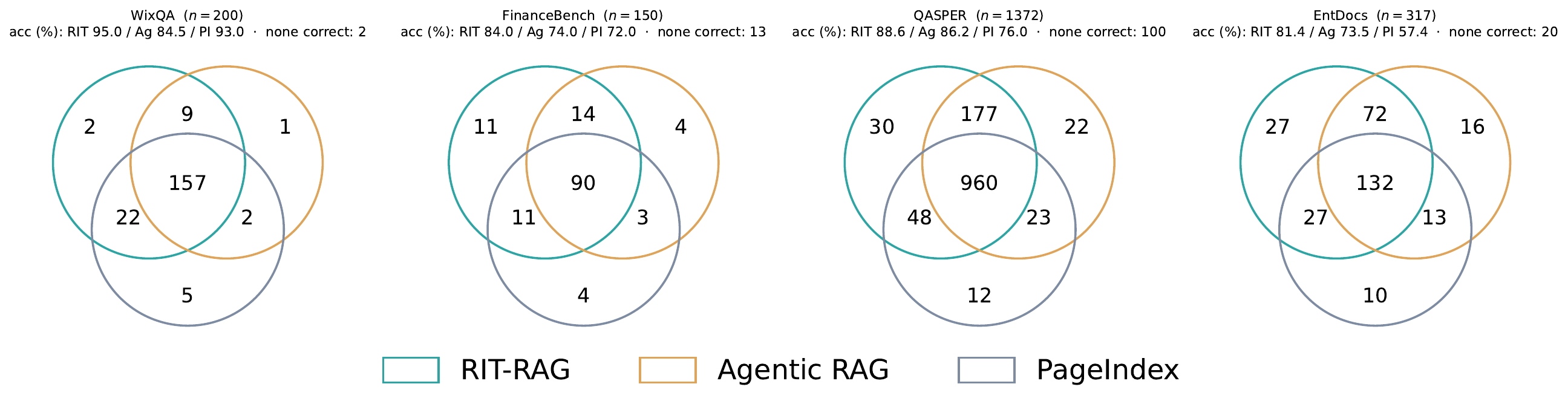}
\caption{Which questions each method answers correctly (Haiku generation), as an
outline Venn. Each colored ring is one method's set of correct questions and region
labels are question counts; ``none correct'' (all three wrong) is noted per panel.}
\label{fig:failure-venn-outline}
\end{figure*}

We compare the questions answered correctly by \ourmethod, PageIndex, and
Agentic RAG. PageIndex and Agentic RAG have complementary strengths: PageIndex
navigates document structure, whereas Agentic RAG searches broadly over
content. Figure~\ref{fig:failure-venn-outline} shows how their correct answers
overlap with those of \ourmethod.

% ---------------------------------------------------------------------------
% Outline-only 3-set Venn (alternative to the UpSet plot, fig:failure-upset):
% colored rings, no fills, so there are no blended overlap colors to decode.
% Generated by analysis/failure_venn_matrix.py.
% ---------------------------------------------------------------------------

% ---------------- Figure: Lockheed Martin (multi-document) ----------------
\ifcsname trStep\endcsname\else
  \PackageError{trace-figures}{trace_macros.tex not loaded. Add \string at the END of preamble.tex}{}
\fi
\begin{figure*}[t]
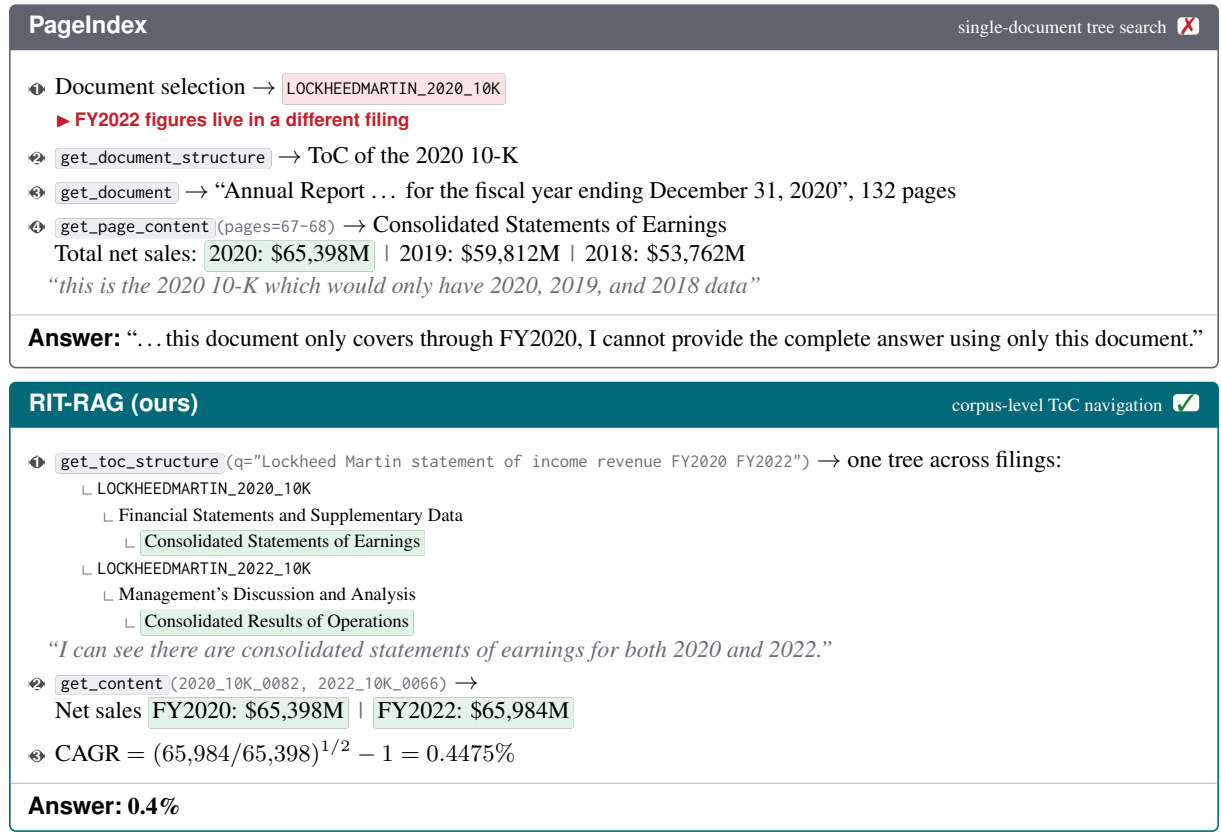

\centering
\begin{qstrip}
\trQ What is Lockheed Martin's 2-year total revenue CAGR from FY2020 to FY2022 (in percent, one decimal)? Use the statement of income.
\par\smallskip\trGold \textbf{0.4\%}
\end{qstrip}
\vspace{-3pt}
\begin{tcbraster}[raster columns=1, raster row skip=5pt]
\begin{tracebox}[base,badans]{PageIndex}{\trBAD}{single-document tree search}
\trStep{Document selection $\rightarrow$ \trBad{\trDoc{LOCKHEEDMARTIN\_2020\_10K}} \\
  \trWhy{FY2022 figures live in a different filing}}
\trStep{\trTool{get\_document\_structure} $\rightarrow$ ToC of the 2020 10-K}
\trStep{\trTool{get\_document} $\rightarrow$ ``Annual Report \ldots\ for the fiscal year ending December 31, 2020'', 132 pages}
\trStep{\trTool{get\_page\_content}\trArgs{pages=67-68} $\rightarrow$ Consolidated Statements of Earnings \\
  Total net sales: \trGood{2020: \$65,398M} \trSep{} 2019: \$59,812M \trSep{} 2018: \$53,762M}
\trSub{\trThought{this is the 2020 10-K which would only have 2020, 2019, and 2018 data}}
\tcblower
\trAns ``\ldots this document only covers through FY2020, I cannot provide the complete answer using only this document.''
\end{tracebox}
\begin{tracebox}[ours,okans]{\ourmethod{} (ours)}{\trOK}{corpus-level ToC navigation}
\trStep{\trTool{get\_toc\_structure}\trArgs{q="Lockheed Martin statement of income revenue FY2020 FY2022"} $\rightarrow$ one tree across filings:
  \trToc{0}{\trDoc{LOCKHEEDMARTIN\_2020\_10K}}
  \trToc{1}{Financial Statements and Supplementary Data}
  \trToc{2}{\trGood{Consolidated Statements of Earnings}}
  \trToc{0}{\trDoc{LOCKHEEDMARTIN\_2022\_10K}}
  \trToc{1}{Management's Discussion and Analysis}
  \trToc{2}{\trGood{Consolidated Results of Operations}}}
\trSub{\trThought{I can see there are consolidated statements of earnings for both 2020 and 2022.}}
\trStep{\trTool{get\_content}\trArgs{2020\_10K\_0082, 2022\_10K\_0066} $\rightarrow$ \\
  Net sales \trGood{FY2020: \$65,398M} \trSep{} \trGood{FY2022: \$65,984M}}
\trStep{CAGR $= (65{,}984 / 65{,}398)^{1/2} - 1 = 0.4475\%$}
\tcblower
\trAns \textbf{0.4\%}
\end{tracebox}
\end{tcbraster}
\vspace{2pt}
\caption{\textbf{Multi-document question.} PageIndex commits to a single filing before searching and can never see the FY2022 revenue. \ourmethod{} retrieves a corpus-level table of contents, locates the income statement in \emph{both} 10-Ks, and answers in two tool calls. Traces abridged (\trElide); quoted text is verbatim model output.}
\label{fig:trace_lockheed}
\end{figure*}

% ---------------- Figure: Block FY2016 working capital (structure fragmentation) ----------------
% Source: FB qwen prep, Haiku gen, results.jsonl line 57 (see PROVENANCE.md).
\ifcsname trStep\endcsname\else
  \PackageError{trace-figures}{trace_macros.tex not loaded. Add \string at the END of preamble.tex}{}
\fi
\begin{figure*}[t]
\centering
\begin{qstrip}
\trQ What is Block's FY2016 working capital ratio (total current assets / total current liabilities)?
\par\smallskip\trGold \textbf{1.73}
\end{qstrip}
\vspace{-3pt}
\begin{tcbraster}[raster columns=1, raster row skip=5pt]
\begin{tracebox}[base,badans]{Agentic RAG}{\trBAD}{flat chunk search}
\trStep{\trTool{search\_corpus} $\times 10$ $\rightarrow$ chunk of the 2016 balance sheet:\\[2pt]
  {\scriptsize\setlength{\tabcolsep}{4pt}%
  \begin{tabular}{@{}lr@{}}
  \toprule
  Total current assets & 1{,}001{,}425 \\
  Accounts payable & 12{,}602 \\
  Customers payable & 388{,}058 \\
  \multicolumn{2}{@{}l}{\trBad{\itshape chunk ends here, mid-table}} \\ \bottomrule
  \end{tabular}}}
\trSub{\trWhy{the total current liabilities line falls outside the chunk}}
\trStep{Adds \trBad{22{,}472} from a note table in the \emph{2017} 10-K}
\trStep{Current liabilities $= 12{,}602 + 388{,}058 + 22{,}472 = \trBad{423{,}132}$}
\trSub{Ratio $= 1{,}001{,}425 / 423{,}132$}
\tcblower
\trAns \textbf{2.37}
\end{tracebox}
\begin{tracebox}[ours,okans]{\ourmethod{} (ours)}{\trOK}{corpus-level ToC navigation}
\trStep{\trTool{get\_toc\_structure} $\rightarrow$ \trGood{Consolidated Balance Sheets} node of the FY2016 filing}
\trStep{\trTool{get\_content} $\rightarrow$ the whole statement as one node:\\[2pt]
  {\scriptsize\setlength{\tabcolsep}{4pt}%
  \begin{tabular}{@{}lr@{}}
  \toprule
  Total current assets & 1{,}001{,}425 \\
  \multicolumn{2}{@{}l}{\trElide four liability rows \trElide} \\
  Total current liabilities & \trGood{577{,}464} \\ \bottomrule
  \end{tabular}}}
\trSub{\trWhyOK{the node boundary is the statement boundary, so no row is cut off}}
\trStep{Ratio $= 1{,}001{,}425 / 577{,}464$}
\tcblower
\trAns \textbf{1.73}
\end{tracebox}
\end{tcbraster}
\vspace{2pt}
\caption{\textbf{Structure fragmentation.} Agentic RAG retrieves the right balance sheet, but its fixed-size chunk ends mid-table, before the total-current-liabilities line. After ten searches it sums the visible rows plus a figure from a note in a \emph{different} filing, and answers 2.37. \ourmethod{} navigates to the Consolidated Balance Sheets node and reads the full statement, so the total is available and the ratio is correct. Traces abridged (\trElide).}
\Description{Two stacked traces. Agentic RAG retrieves a chunk of Block's 2016 balance sheet that ends after the customers-payable row, adds 22,472 from a 2017 10-K note, and answers 2.37. RIT-RAG reads the Consolidated Balance Sheets node, finds total current liabilities of 577,464, and answers 1.73.}
\label{fig:trace_block}
\end{figure*}

Among the questions answered correctly by PageIndex but not by Agentic RAG,
\ourmethod answers 22 of 27 on WixQA, 11 of 15 on FinanceBench, 48 of 60 on
QASPER, and 27 of 37 on \entdocs(Figure~\ref{fig:trace_block}). Among those answered correctly by Agentic RAG but not by PageIndex,
\ourmethod answers 9 of 10, 14 of 18, 177 of 199, and 72 of 88, respectively (Figure~\ref{fig:trace_lockheed}). It also
answers 2, 11, 30, and 27 questions that both baselines miss. These results show
that \ourmethod retains most of the distinct strengths of both baselines while
also solving some questions that neither baseline answers correctly.

\section{Why graph baselines fail on QASPER and FinanceBench}
\label{app:graph-failure}
\begin{figure*}[t]
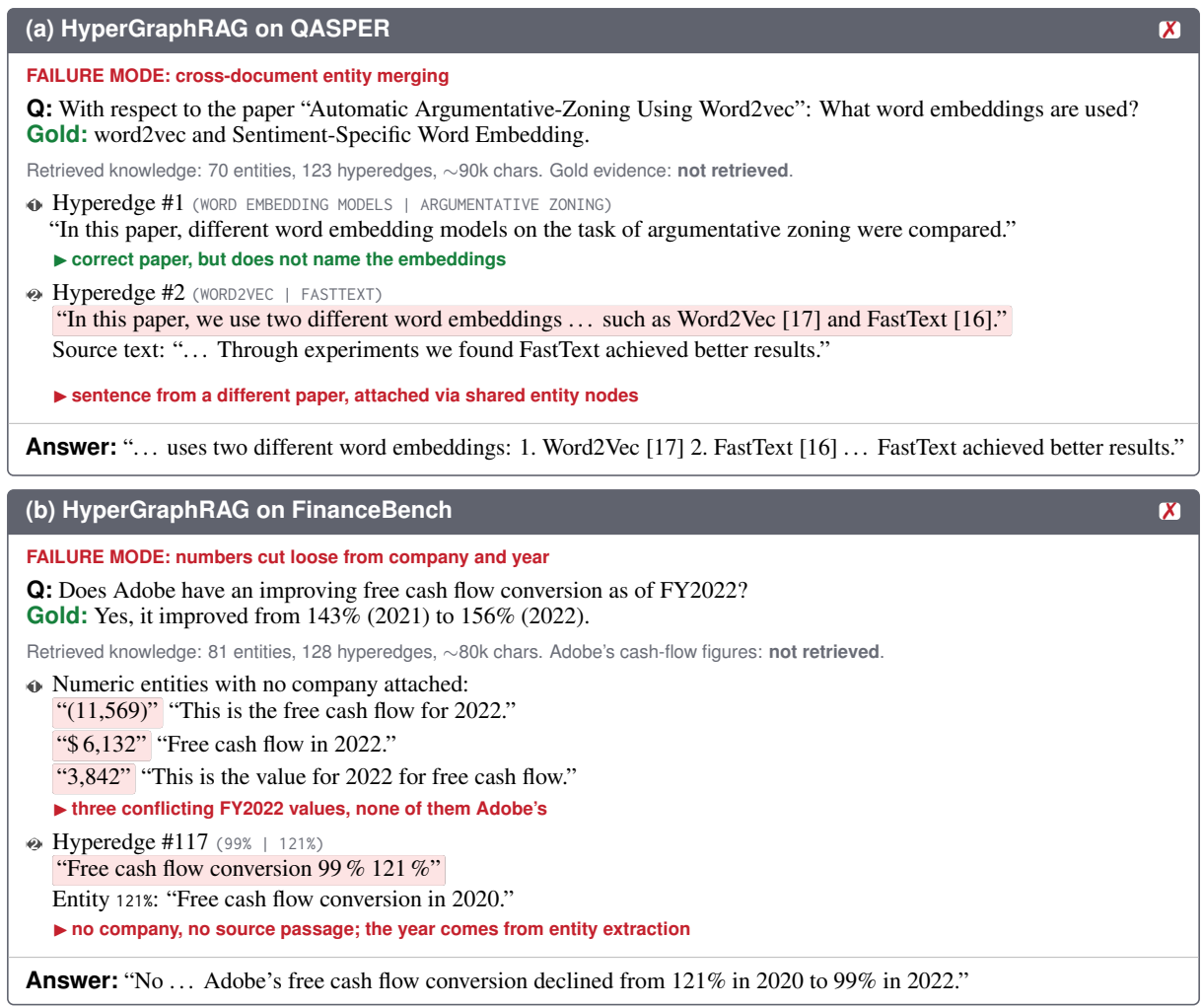

\centering
\begin{tcbraster}[raster columns=1, raster row skip=5pt]
\begin{tracebox}[base,badans]{(a) HyperGraphRAG on QASPER}{\trBAD}{}
{\color{badC}\sffamily\bfseries\scriptsize FAILURE MODE: cross-document entity merging}\par\smallskip
\trQ With respect to the paper ``Automatic Argumentative-Zoning Using Word2vec'': What word embeddings are used?\\
\trGold word2vec and Sentiment-Specific Word Embedding.
\par\smallskip{\color{mutedC}\scriptsize\sffamily Retrieved knowledge: 70 entities, 123 hyperedges, $\sim$90k chars. Gold evidence: \textbf{not retrieved}.}
\trStep{Hyperedge \#1 \trArgs{WORD EMBEDDING MODELS \textbar{} ARGUMENTATIVE ZONING}\\
  ``In this paper, different word embedding models on the task of argumentative zoning were compared.''}
\trSub{\trWhyOK{correct paper, but does not name the embeddings}}
\trStep{Hyperedge \#2 \trArgs{WORD2VEC \textbar{} FASTTEXT}\\
  \trBad{``In this paper, we use two different word embeddings \ldots\ such as Word2Vec [17] and FastText [16].''}}
\trSub{Source text: ``\ldots\ Through experiments we found FastText achieved better results.''}
\trSub{\trWhy{sentence from a different paper, attached via shared entity nodes}}
\tcblower
\trAns ``\ldots\ uses two different word embeddings: 1.~Word2Vec [17] 2.~FastText [16] \ldots\ FastText achieved better results.''
\end{tracebox}
\begin{tracebox}[base,badans]{(b) HyperGraphRAG on FinanceBench}{\trBAD}{}
{\color{badC}\sffamily\bfseries\scriptsize FAILURE MODE: numbers cut loose from company and year}\par\smallskip
\trQ Does Adobe have an improving free cash flow conversion as of FY2022?\\
\trGold Yes, it improved from 143\% (2021) to 156\% (2022).
\par\smallskip{\color{mutedC}\scriptsize\sffamily Retrieved knowledge: 81 entities, 128 hyperedges, $\sim$80k chars. Adobe's cash-flow figures: \textbf{not retrieved}.}
\trStep{Numeric entities with no company attached:\\
  \trBad{``(11,569)''}~``This is the free cash flow for 2022.''\\
  \trBad{``\$\,6,132''}~``Free cash flow in 2022.''\\
  \trBad{``3,842''}~``This is the value for 2022 for free cash flow.''}
\trSub{\trWhy{three conflicting FY2022 values, none of them Adobe's}}
\trStep{Hyperedge \#117 \trArgs{99\% \textbar{} 121\%}\\
  \trBad{``Free cash flow conversion 99\,\% 121\,\%''}\\
  Entity \trDoc{121\%}: ``Free cash flow conversion in 2020.''}
\trSub{\trWhy{no company, no source passage; the year comes from entity extraction}}
\tcblower
\trAns ``No \ldots\ Adobe's free cash flow conversion declined from 121\% in 2020 to 99\% in 2022.''
\end{tracebox}
\end{tcbraster}
\vspace{2pt}
\caption{\textbf{Failure modes of graph-based retrieval.} (a)~Entity nodes such as \textsc{Word2Vec} are shared across the whole corpus, so a hyperedge from an unrelated paper is retrieved as evidence and the model reports FastText. (b)~Table cells are extracted as free-floating numeric entities without their company or fiscal period. The retrieved subgraph holds several contradictory ``FY2022 free cash flow'' values, and the answer is built on a conversion ratio that does not belong to Adobe. In both cases the gold evidence is absent from $\sim$80--90k characters of retrieved knowledge.}
\label{fig:trace-graph-failures}
\end{figure*}

The graph baselines are strong on WixQA yet collapse on QASPER and FinanceBench, for
two different reasons. Figure~\ref{fig:trace-graph-failures} gives representative
traces from both datasets.

On QASPER the cause is a loss of document scoping. QASPER is per-paper question
answering, but HyperGraphRAG builds a single graph over all $1{,}585$ papers with
entities keyed by surface form, so the generic vocabulary of scientific writing fuses
into corpus-spanning hubs (one \texttt{DATASET} node linked to $1{,}728$ passages,
\texttt{MODEL} $4{,}046$, \texttt{BERT} $1{,}397$). A question about one paper then
retrieves fragments pooled from unrelated papers, and the model copies the wrong
paper's facts (Figure~\ref{fig:trace-graph-failures}\emph{(a)}).

On FinanceBench the answers live in tables, but graph extraction shreds each table into
fragments and stores bare figures as entities (about a fifth of all nodes are pure
numbers, which are poor retrieval keys), so the gold evidence is rarely surfaced,
resulting in hallucinations (Figure~\ref{fig:trace-graph-failures}\emph{(b)}).

% fig:topk belongs to the Sensitivity section below. It is declared here, ahead of
% the full-width trace figure, so LaTeX can place it beside the last text column
% instead of leaving it for a nearly empty page at the end.
% Sensitivity of RIT-RAG to n, the number of nodes kept by the Search tool (k' = 4n chunks). 26 Sep.
\begin{figure}[t]
\centering
\includegraphics[width=\linewidth]{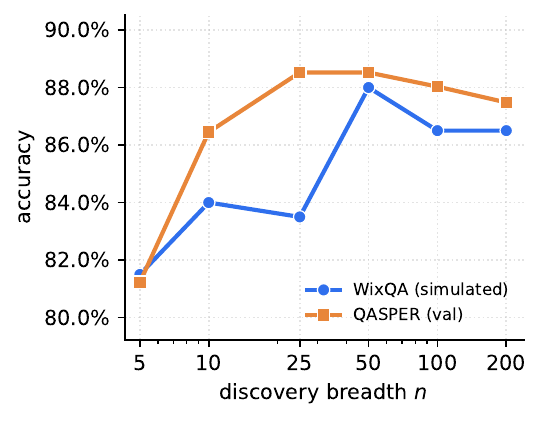}
\caption{Accuracy of \ourmethod against the number of nodes $\numtopnodes$ that the \gettoc{} tool keeps, on held-out questions of WixQA (simulated split) and QASPER (validation split), with \haiku for preprocessing and generation.}
\label{fig:topk}
\end{figure}

% fig:trace-graph-failures is input after \onecolumn below: declared here it was flushed by
% \onecolumn onto a page of its own (3/4 empty); there it shares the first prompts page.

% ===========================================================================
\section{\ourmethod Behavior Statistics}
\label{app:tool-calls}

Table~\ref{tab:tool-calls} reports per-query tool-call statistics for \ourmethod
on all four benchmarks under \haiku generation. \ourmethod navigates with two tools, \gettoc{} and \readtool{}.

\begin{table*}[t]
\centering
\small
\setlength{\tabcolsep}{6pt}
\caption{Per-query tool-call statistics for \ourmethod (\haiku generation).
\textbf{ToC}/\textbf{Cont.}: mean calls to
\texttt{get\_toc\_structure}/\texttt{get\_content};
\textbf{Total}: their sum;
\textbf{Nodes}: mean content nodes per query;
\textbf{N/call}: nodes per \texttt{get\_content} call.}
\label{tab:tool-calls}
\begin{tabular}{l r r r r r r}
\toprule
& & \multicolumn{3}{c}{Mean calls per query} & & \\
\cmidrule(lr){3-5}
Benchmark & $n$ & Total & ToC & Cont. & Nodes & N/call \\
\midrule
WixQA         & 200  & 2.99 & 1.33 & 1.66 & 5.5 & 3.3 \\
QASPER        & 1372 & 2.75 & 1.43 & 1.32 & 3.3 & 2.5 \\
FinanceBench  & 150  & 2.70 & 1.46 & 1.24 & 2.7 & 2.2 \\
\entdocsbench & 317  & 2.99 & 1.47 & 1.51 & 3.8 & 2.5 \\
\bottomrule
\end{tabular}
\end{table*}

Tool calls made by \ourmethod ranges from $2.70$ to $2.99$ per query. The agent chooses to spend similar amount of calls for both the
navigation and reading tools. \gettoc{} calls stay near $1.3$--$1.5$ and \readtool{} calls near
$1.2$--$1.7$ on every benchmark, so corpus size does not add agent iterations. \ourmethod carefully chooses what to read, reading only $2.7$--$5.5$ nodes and stops early if it has found the answer to the question. The agent chooses to request about 2-3 node's content in a single \readtool{}, thereby reading the corpus content at once instead of wasting multiple tool calls by just by reading a single node.

% ===== BEGIN: Sensitivity to n (added 26 Sep, Yatin) =====
\section{Sensitivity to the Number of Nodes}
\label{app:ablations}

The Search tool retrieves the top-$k' = 4\numtopnodes$ chunks, and keeps the $\numtopnodes$ nodes with the highest scores to induce the subtree (Section~\ref{sec:ourmethod}).
A small $\numtopnodes$ may miss the relevant nodes, whereas a large $\numtopnodes$ adds more nodes to the context of the agent.
We choose $\numtopnodes$ on held-out questions, \viz, the simulated split of WixQA ($200$ questions) and the validation split of QASPER ($819$ questions), with \haiku for preprocessing and generation (Figure~\ref{fig:topk}).
The accuracy increases with $\numtopnodes$ up to $\numtopnodes = 50$, and then decreases slightly.
WixQA reaches its best accuracy of $88.0\%$ at $\numtopnodes = 50$, and QASPER reaches $88.5\%$ at both $\numtopnodes = 25$ and $50$.
We therefore use $\numtopnodes = 50$ in all our experiments.
% ===== END: Sensitivity to n =====

% ===========================================================================
\section{\entdocsbench Creation}
\label{sec:entdocs-creation}
The \entdocsbench is designed to evaluate agents' ability to perform retrieval on large corpora. Specifically, the \entdocs are scraped from IBM Documentation\footnote{\url{https://www.ibm.com/docs/en/products}}, amounting to $\sim$3M documents (9M chunks) spanning dense technical topics such as compilers for the Z product line, COBOL troubleshooting, and various other documentation for IBM Products.

To construct the question bank, we follow the data generation procedure described in KARL \cite{sun2026karl}. A few hand-written questions are provided to an agent equipped with a search tool over the corpus as few-shot examples. The agent is then allowed to explore the corpus and formulate questions grounded in the retrieved documents. Each question is generated with citations tracing it back to the source documents, allowing us to retain provenance and evaluate retrieval performance.

We use Claude Sonnet 5 as the question generation agent and \texttt{text-embedding-3-small} \cite{openai_embeddings} to index the corpus chunks. We follow the prompts and filtering procedure from KARL \cite{sun2026karl}. After filtering, the final benchmark contains 317 questions spanning 350 webpages. The questions cover three distinct topologies, ranging from single-webpage questions to multi-step questions requiring evidence from 3--4 webpages. Each question is accompanied by a gold answer, supporting citations, and a set of answer nuggets used to assess answer completeness.

\section{LLM Usage}

\phead{Writing support}
During the preparation of this manuscript, we employed a Large Language Model (LLM) as a writing support tool. Specifically, the LLM was used to polish the phrasing, improve grammatical accuracy, and provide paraphrased alternatives to enhance clarity and readability. The LLM's role was limited to language refinement, and all suggested edits were reviewed and verified by the authors before inclusion.

% \phead{Synthetic benchmark generation}
% We also used an LLM to construct the IBM~Z evaluation benchmark, since no
% question-answering benchmark exists for this corpus. The LLM was prompted with
% content sampled from the crawled documentation pages and asked to produce
% question-answer pairs grounded in that content, together with a supporting
% reasoning chain and citations to the specific source pages used. The resulting
% questions span both single-page and multi-page reasoning and a range of reasoning
% patterns (for example, causal and comparative). Each answer is grounded in the
% cited pages, and the gold evidence is recorded at page (node) granularity to match
% the retrieval metrics. We then automatically audited the generated set with a
% separate LLM judge (\texttt{gpt-5-chat}), which scored answer grounding and the
% validity of each cited URL; questions whose citations pointed to stale or
% uncrawled pages were corrected or removed. The authors manually reviewed the final
% $317$-question set. The LLM was used only to generate and audit candidate
% questions; it was never used to produce or influence the retrieval or
% answer-generation results reported for any method.

% ===========================================================================
% Single-column layout from here to the end: the prompts only, so they are
% copy-pastable. Everything before this is two-column. \onecolumn starts a new page.
\onecolumn
% LaTeX keeps the previous two-column page's reduced column height here (that
% column held a float), which made the first prompt box break early. Reset it.
\makeatletter\global\@colroom\@colht \global\vsize\@colht\makeatother
% ---------------- Figure: graph-RAG failure modes ----------------
\ifcsname trStep\endcsname\else
  \PackageError{trace-figures}{trace_macros.tex not loaded. Add \string at the END of preamble.tex}{}
\fi

\section{Prompts}
\label{app:prompts}

\subsection{\ourmethod System Prompt}
\label{app:agent-prompt}

\begin{promptbox}
\begin{lstlisting}
You are a document QA assistant.

You have two tools available:
- get_toc_structure: Uses hybrid dense+sparse search to return a table of contents
  -- a hierarchy of nodes relevant to the query. Each node has "id", "title",
  "is_readable" (whether the node's content can be fetched),
  "metadata" (dataset-specific fields), and child "nodes". You may call it
  multiple times with different queries to explore different sections.
- get_content: Fetches the full text content for a list of node IDs.

STRATEGY:
1. Before every tool call, write a short reasoning block explaining:
   - What you already know from previous results (or "nothing yet" on the first call)
   - Why the previous results were insufficient (if applicable)
   - What specific query or node IDs you will use next and why
2. Call get_toc_structure with a query. Inspect the returned nodes to identify
   the most promising ones.
3. Call get_content with the most promising node IDs -- request several at once
   (aim for 3-5 in a single call), not one at a time.
4. If the content is insufficient, either (a) call get_toc_structure again with a
   refined query to discover more relevant nodes, or (b) call get_content with
   additional readable node IDs from the current TOC. Repeat as needed.
5. When you have enough information, respond in this exact format:
<reasoning>your step-by-step reasoning</reasoning>
<answer>the answer only, no preamble</answer>
<sources>node_ids you used, e.g. 1, 2</sources>

RULES:
- Always write your reasoning BEFORE every tool call -- never skip it.
- Only pass node IDs where "is_readable": true to get_content. Nodes with
  "is_readable": false exist only to show hierarchy -- they have no fetchable content.
- Do NOT call get_toc_structure and get_content in the same turn.
- MANDATORY: You must call get_content at least once and base your answer ONLY on
  the full text it returns. You cannot answer -- or conclude the answer is not in
  the corpus -- without reading content first.
- Titles and any summaries/keywords are navigation hints ONLY. They are lossy and
  never sufficient to answer: a relevant-looking title or summary means you should
  FETCH that node with get_content, not that you may answer from it. Do not guess
  or use prior knowledge.
- Be concise -- include only crucial information relevant to the question.
- Budget: reading content with get_content is what answers the question --
  get_toc_structure only points at nodes and never answers anything, so do not keep
  re-searching when you already have promising leads; fetch and read them instead.
  You have at most 10 tool calls total (at most 3 of them get_toc_structure calls);
  spend the rest reading content. Before each turn you receive a [CONTEXT USAGE]
  message; on a [CONTEXT WARNING] (>=80% used), stop searching and answer from the
  content you already have.

NODE METADATA: Nodes include additional fields:
- "summary": short description of the node's content -- use it to judge relevance
  before calling get_content
- "keywords": list of key terms in this node -- match against the question to
  identify relevant nodes
\end{lstlisting}
\end{promptbox}

\subsection{WixQA Judge Prompt}

\begin{promptbox}
\begin{lstlisting}
You are an expert evaluator for AI-generated answers to questions.
Your task is to determine whether the AI-generated answer correctly answers the
question. The golden answers below are all valid correct answers -- the AI answer
is correct if it matches any one of them.

Evaluation Criteria:
- The AI-generated answer is correct if it captures the essential meaning or key
  facts of any golden answer, even if phrased differently.
- If the AI-generated answer is a superset of a golden answer (more detailed but not
  factually wrong), it is correct.
- The AI-generated answer is correct if it conveys the same conclusion or equivalent
  information as any golden answer.
- A vague or generic answer that does not actually address the specific question is
  incorrect.
- The AI-generated answer is incorrect if it contradicts key facts in all golden
  answers or answers a different question entirely.

Inputs:
- Question: {question}
- AI-Generated Answer: {predicted_answer}
- Golden Answers (any one match is sufficient):
{gold_answers}

Decide whether the AI-generated answer is correct.
\end{lstlisting}
\end{promptbox}

\subsection{FinanceBench Judge Prompt}

\begin{promptbox}
\begin{lstlisting}
You are an expert evaluator for AI-generated responses to queries. Your task is to
determine whether the AI-generated answer correctly answers the query based on the
golden answer provided by a human expert.

Numerical Accuracy:
- Rounding differences should be ignored if they do not meaningfully change the
  conclusion.
- You can allow some flexibility in accuracy. For example, 1.2 is considered similar
  to 1.23. Two numbers are considered similar if one can be rounded to the other.
- Fractions, percentage, and numerics could be considered similar, for example:
  "11 of 14" is considered equivalent to "79%" and "0.79".

Evaluation Criteria:
- If the golden answer or any of its equivalence can be inferred or generated from
  the AI-generated answer, then the AI-generated answer is considered correct.
- If any number, percentage, fraction, or figure in the golden answer is not present
  in the AI-generated answer, but can be inferred or generated from the AI-generated
  answer or implicitly exist in the AI-generated answer, then the AI-generated answer
  is considered correct.
- The AI-generated answer is considered correct if it conveys the same or similar
  meaning, conclusion, or rationale as the golden answer.
- If the AI-generated answer is a superset of the golden answer, it is also correct.
- If the AI-generated answer provides a valid answer or reasonable interpretation
  compared to the golden answer, it is considered correct.
- If the AI-generated answer contains subjective judgments or opinions, it is
  considered correct as long as they are reasonable and justifiable compared to the
  golden answer.
- Otherwise, the AI-generated answer is incorrect.

Inputs:
- Question: {question}
- AI-Generated Answer: {predicted_answer}
- Golden Answer: {gold_answer}

Your output should be ONLY a boolean value: `True` or `False`, nothing else.
\end{lstlisting}
\end{promptbox}

\subsection{\entdocs Judge Prompt}

\begin{promptbox}
\begin{lstlisting}
You are an expert evaluator for AI-generated responses to IBM technical documentation
queries. Your task is to determine whether the AI-generated answer correctly answers
the query based on the golden answer provided by a human expert.

Evaluation Criteria:
- The AI-generated answer is correct if it captures the essential meaning, key steps,
  or core concepts of the golden answer, even if phrased differently or in a
  different order.
- If the golden answer is a list of components or steps, the AI-generated answer is
  correct if it covers the most important ones. Minor omissions of secondary details
  are acceptable.
- If the AI-generated answer is a superset of the golden answer (more detailed but
  not wrong), it is correct.
- If the AI-generated answer uses different but equivalent technical terminology for
  the same IBM concept, tool, or process, it is correct.
- The AI-generated answer is incorrect if it contradicts the golden answer on a key
  fact, names the wrong tool/product/command, describes the wrong process, or answers
  a different question entirely.
- A vague or generic answer that does not actually address the specific question is
  incorrect.

Inputs:
- Question: {question}
- AI-Generated Answer: {predicted_answer}
- Golden Answer: {gold_answer}

Your output should be ONLY a boolean value: `True` or `False`, nothing else.
\end{lstlisting}
\end{promptbox}

\subsection{G-E Judge Prompts (Seven Dimensions)}
\label{app:ge-prompts}
The graded G-E judge follows HyperGraphRAG~\citep{luo2025hypergraphrag}: one call per
dimension scores the answer on a $0$--$10$ integer scale, read from a structured
\texttt{\{"score": <int>\}} output. We report all seven dimensions (correctness,
relevance, factuality, comprehensiveness, knowledgeability, logical coherence, and
diversity). Every dimension uses the shared template below, with the dimension-specific
\texttt{title}, \texttt{goal}, and scoring \texttt{rubric} substituted from the list that
follows.
\begin{promptbox}
\begin{lstlisting}
---Role--
You are a helpful assistant evaluating the **{title}** of a generated response.

---Question--
{question}

---Golden Answers--
{gold_answers}

---Evaluation Goal--
Evaluate **{goal}** using a **0-10 integer scale**. {rubric}

---Generation to be Evaluated--
{generation}

Respond ONLY with JSON: {"score": <integer 0-10>}
\end{lstlisting}
\end{promptbox}

The seven \texttt{(title, goal, rubric)} substitutions are:

\begin{promptbox}
\begin{lstlisting}
[correctness]
goal: whether the reasoning and answer are logically and factually correct
- 10: Fully accurate and logically sound; no flaws in reasoning or facts.
- 8-9: Mostly correct with minor inaccuracies or small logical gaps.
- 6-7: Partially correct; some key flaws or inconsistencies present.
- 4-5: Noticeable incorrect reasoning or factual errors throughout.
- 1-3: Largely incorrect, misleading, or illogical.
- 0: Entirely wrong or nonsensical.

[relevance]
goal: whether the reasoning and answer are highly relevant and helpful to the question
- 10: Fully focused on the question; highly relevant and helpful.
- 8-9: Mostly on point; minor digressions but overall useful.
- 6-7: Generally relevant, but includes distractions or less helpful parts.
- 4-5: Limited relevance; much of the response is off-topic or unhelpful.
- 1-3: Barely related to the question or largely unhelpful.
- 0: Entirely irrelevant.

[factuality]
goal: whether the reasoning and answer are based on accurate and verifiable facts
- 10: All facts are accurate and verifiable.
- 8-9: Mostly accurate; only minor factual issues.
- 6-7: Contains some factual inaccuracies or unverified claims.
- 4-5: Several significant factual errors.
- 1-3: Mostly false or misleading.
- 0: Completely fabricated or factually wrong throughout.

[comprehensiveness]
goal: whether the thinking considers all important aspects and is thorough
- 10: Extremely thorough, covering all relevant angles and considerations with depth.
- 8-9: Covers most key aspects clearly and thoughtfully; only minor omissions.
- 6-7: Covers some important aspects, but lacks depth or overlooks notable areas.
- 4-5: Touches on a few relevant points, but overall lacks substance or completeness.
- 1-3: Sparse or shallow treatment of the topic; misses most key aspects.
- 0: No comprehensiveness at all; completely superficial or irrelevant.

[knowledgeability]
goal: whether the thinking is rich in insightful, domain-relevant knowledge
- 10: Demonstrates exceptional depth and insight with strong domain-specific knowledge.
- 8-9: Shows clear domain knowledge with good insight; mostly accurate and relevant.
- 6-7: Displays some understanding, but lacks depth or has notable gaps.
- 4-5: Limited knowledge shown; understanding is basic or somewhat flawed.
- 1-3: Poor grasp of relevant knowledge; superficial or mostly incorrect.
- 0: No evidence of meaningful knowledge.

[logical coherence]
goal: whether the reasoning is internally consistent, clear, and well-structured
- 10: Highly logical, clear, and easy to follow throughout.
- 8-9: Well-structured with minor lapses in flow or clarity.
- 6-7: Some structure and logic, but a few confusing or weakly connected parts.
- 4-5: Often disorganized or unclear; logic is hard to follow.
- 1-3: Poorly structured and incoherent.
- 0: Entirely illogical or unreadable.

[diversity]
goal: whether the reasoning is thought-provoking, offering varied or novel perspectives
- 10: Exceptionally rich and original; demonstrates multiple fresh and thought-provoking ideas.
- 8-9: Contains a few novel angles or interesting perspectives.
- 6-7: Some variety, but generally safe or conventional.
- 4-5: Mostly standard thinking; minimal diversity.
- 1-3: Very predictable or monotonous.
- 0: No diversity or originality at all.
\end{lstlisting}
\end{promptbox}

\end{document}